\documentclass[10pt,twocolumn,letterpaper]{article}

\usepackage{cvpr}              

\usepackage{graphicx}
\usepackage{amsmath}
\usepackage{amssymb}
\usepackage{amsthm}
\usepackage{booktabs}
\usepackage{multirow}
\usepackage{color,xcolor}
\usepackage{soul}
\usepackage{xr-hyper}
\usepackage[accsupp]{axessibility}

\usepackage{amsmath}
\newtheorem{theorem}{Theorem}

\usepackage{floatrow}
\allowdisplaybreaks

\newfloatcommand{figurebox}{figure}[\nocapbeside][\dimexpr(\textwidth-\columnsep)/2\relax]
\newfloatcommand{tablebox}{table}[\nocapbeside][\dimexpr(\textwidth-\columnsep)/2\relax]

\usepackage[pagebackref,breaklinks,colorlinks]{hyperref}

\newcommand \footnoteONLYtext[1]
{
	\let \mybackup \thefootnote
	\let \thefootnote \relax
	\footnotetext{#1}
	\let \thefootnote \mybackup
	\let \mybackup \imareallyundefinedcommand
}

\usepackage[capitalize]{cleveref}
\crefname{section}{Sec.}{Secs.}
\Crefname{section}{Section}{Sections}
\Crefname{table}{Table}{Tables}
\crefname{table}{Tab.}{Tabs.}

\def\cvprPaperID{9832} 
\def\confName{CVPR}
\def\confYear{2022}

\begin{document}
\twocolumn
\title{Revisiting the Transferability of Supervised Pretraining: an MLP Perspective}

\author{Yizhou Wang$^{1,3\ast\dag}$, Shixiang Tang$^{2\dag}$, Feng Zhu$^{3}$, Lei Bai$^{2\ddag}$, Rui Zhao$^{3,4}$, Donglian Qi$^1$, Wanli Ouyang$^2$\\
$^1$Zhejiang University, $^2$The University of Sydney,$^3$SenseTime Research,\\ $^4$Qing Yuan Research Institute, Shanghai Jiao Tong University, Shanghai, China\\
{\tt\small \{yizhouwang, qidl\}@zju.edu.cn, stan3906@uni.sydney.edu.au, zhufeng@sensetime.com,}\\
{\tt\small baisanshi@gmail.com, zhaorui@sensetime.com, wanli.ouyang@sydney.edu.au}
}

\maketitle

\begin{abstract}
The pretrain-finetune paradigm is a classical pipeline in visual learning. Recent progress on unsupervised pretraining methods shows superior transfer performance to their supervised counterparts. This paper revisits this phenomenon and sheds new light on understanding the transferability gap between unsupervised and supervised pretraining from a multilayer perceptron (MLP) perspective. While previous works~\cite{chen2020a,chen2020improved,grill2020bootstrap} focus on the effectiveness of MLP on unsupervised image classification where pretraining and evaluation are conducted on the same dataset, we reveal that the MLP projector is also the key factor to better transferability of unsupervised pretraining methods than supervised pretraining methods. Based on this observation, we attempt to close the transferability gap between supervised and unsupervised pretraining by adding an MLP projector before the classifier in supervised pretraining. Our analysis indicates that the MLP projector can help retain intra-class variation of visual features, decrease the feature distribution distance between pretraining and evaluation datasets, and reduce feature redundancy. Extensive experiments on public benchmarks demonstrate that the added MLP projector significantly boosts the transferability of supervised pretraining, \eg \textbf{+7.2\%} top-1 accuracy on the concept generalization task, \textbf{+5.8\%} top-1 accuracy for linear evaluation on 12-domain classification tasks, and \textbf{+0.8\%} AP on COCO object detection task, making supervised pretraining comparable or even better than unsupervised pretraining. 
\end{abstract}

\section{Introduction}

While Supervised Learning with the cross-entropy loss\footnote{In the paper, we specifically use the notation ``SL'' to indicate the conventional supervised learning with the cross-entropy loss.} (SL) were the de facto pretraining paradigm in computer vision~\cite{he2016deep,tang2021mutual,dosovitskiy2020image,liu2020negative} for a long period, recent unsupervised learning methods~\cite{chen2020a,chen2020improved,he2020momentum,chen2020big,grill2020bootstrap,chen2021exploring,chen2021multisiam,caron2021emerging,dwibedi2021little,caron2020unsupervised,zbontar2021barlow,xie2021propagate} show better transfer learning performance on various visual tasks~\cite{grill2020bootstrap,islam2021a,zhao2021what}. 
This raised the question of why unsupervised pretraining surpasses supervised pretraining even though supervised pretraining uses annotations with rich semantic information.

Several works have attempted to explain the better transferability of unsupervised pretraining than supervised pretraining by the following two reasons: (1) \emph{Learning without semantic information in annotations}~\cite{ericsson2020how,wei2020can,zhao2021what,sariyildiz2021concept}, which makes the backbone less overfit to semantic labels to preserve instance-specific information which may be useful in transfer tasks, and (2) \emph{Special design of the contrastive loss}~\cite{zhao2021what,islam2021a,khosla2020supervised}, which helps the learned features to contain more low/mid-level information for effective transfer to downstream tasks. 
Starting from the perspective of supervision and loss design, these works provide intuitive explanations for better transferability. 

\footnoteONLYtext{$\ast$ The work was done during an internship at SenseTime.}
\footnoteONLYtext{$\dag$ Equal Contribution.}
\footnoteONLYtext{$\ddag$ Corresponding author.}

In this paper, we shed new light on understanding transferability by considering the multilayer perception (MLP) projector. While previous works~\cite{chen2020a,grill2020bootstrap,chen2020improved} verified its effectiveness on the unsupervised image classification task: 
unsupervised training and evaluating the model on the same ImagNet-1K dataset, they did not explore its effectiveness on transfer tasks thoroughly and rigorously. It is not straightforward to extend the effectiveness of MLP on the unsupervised image classification task to downstream tasks if not supported by rigorous experiments or theoretical analysis, because the performance on the pretraining task is not always predictive of the performance on transfer tasks when there exists a large semantic gap~\cite{ericsson2020how,raghu2019transfusion,wang2020comparison}.
To our best knowledge, we are the first to identify the MLP projector as the core factor for the transferability with deep empirical and theoretical analysis. With this new viewpoint, we find that a simple yet effective method, adding an MLP projector, can promote the transferability of the conventional supervised pretraining methods with the cross-entropy loss (SL) to be comparable or even better than representative unsupervised pretraining methods. 

Specifically, 
we use the \emph{concept generalization task}~\cite{sariyildiz2021concept} on ImageNet-1K, where the pretraining and the evaluation datasets have a large semantic distance, as a probe to analyze the transferability of different models. 
Our experimental results and corresponding analysis indicate that the MLP projector in unsupervised pretraining methods is important for their better transferability. Motivated by this observation, we insert an MLP projector before the classifier in SL, forming SL-MLP. The added MLP can improve the transferability of supervised pretraining, making supervised pretraining comparable or even better than unsupervised pretraining. Experimental results on SL and SL-MLP show three interesting findings: 1) The added MLP preserves the intra-class variation on the pretraining dataset. 2) The added MLP decreases the feature distribution distance between the pretraining and the evaluation dataset; 3) The added MLP decreases the feature redundancy in the pretraining dataset. We also provide theoretical analysis on how the preserved intra-class variation and the decreased feature distribution distance improve the performance on the target dataset, by adding an MLP projector. 

Extensive experimental results confirm that adding an MLP projector into the supervised pretraining method (SL) can consistently improve the transferability of the model on various downstream tasks.
Specifically, on the concept generalization task~\cite{sariyildiz2021concept}, SL-MLP boosts the top-1 accuracy compared to SL (55.9\%$\rightarrow$63.1\%) by \textbf{+7.2\%}. It also achieves better performance (64.1\%) than Byol (62.3\%) by \textbf{+1.8\%} on the 300-epochs pretraining setting. 
In classification tasks on 12 cross-domain datasets~\cite{islam2021a}, SL-MLP improves SL by \textbf{+5.8}\% accuracy on average.
Moreover, SL-MLP shows better transferability than SL on COCO object detection~\cite{lin2014microsoft} by \textbf{+0.8}\% AP. These improvements brought by the MLP projector can largely bridge the transferability gap between supervised and unsupervised pretraining as detailed in Sec.~\ref{sec:experimental_results}.

The main contributions of our paper are three-fold. (1) We reveal that the MLP projector is the main factor for the transferability gap between existing unsupervised and supervised learning methods. (2) We empirically demonstrate that, by adding an MLP projector, supervised pretraining methods can have comparable or even better transferability than representative unsupervised pretraining methods. (3) We theoretically prove that the MLP projector can improve transferability of pretrained models by preserving intra-class feature variation.


\section{Related Works}
\noindent\textbf{MLP in unsupervised learning methods.} Adding a multilayer perceptron (MLP) projector after the encoder was first introduced in SimCLR~\cite{chen2020a} and followed by recent unsupervised learning frameworks~\cite{chen2020improved,grill2020bootstrap, caron2020unsupervised, xie2021propagate, chen2021exploring, chen2020big }. SimCLR claims that the MLP can reduce the loss of information caused by the contrastive loss, and various works~\cite{chen2020a,chen2020improved} have verified that the MLP projector can enhance the discriminative ability of unsupervised models on the unsupervised image classification task, where unsupervised training and evaluation are conducted on the same dataset.
However, the relation between the MLP and the transferability of unsupervised learning methods is under-explored. In this paper, we reveal that the MLP projector is also important for the desirable transferability of unsupervised learning.

\noindent\textbf{MLP in supervised learning methods.}
The typical supervised learning method (SL) only uses the cross-entropy loss and shows inferior performance on various transfer tasks than recent unsupervised learning methods. Inspired by~\cite{xie2021detco,dwibedi2021little}, recent works~\cite{khosla2020supervised, islam2021a} introduced the contrastive loss equipped with an MLP projector into SL to improve its transferability. Nonetheless, those works ignored the ablation on the MLP and attributed the better transfer performance to the contrastive mechanism in the loss. In this paper, we propose that the MLP is important for the improved transferability of recent supervised learning methods
~\cite{khosla2020supervised, islam2021a}
with empirical and theoretical analysis.

\noindent\textbf{Transferability gap between supervised and unsupervised learning.} Previous works attributed the superior transferability of unsupervised learning to \textit{lack of annotation}~\cite{sariyildiz2021concept,ericsson2020how,wei2020can,zhao2021what} or \textit{special design of the contrastive loss}~\cite{islam2021a,khosla2020supervised,zhao2021what,tang2021gradient}. Different from both reasons, we explain the transferability gap by considering the architectural difference between the supervised and the unsupervised learning frameworks. From this perspective, we analyze the role of the MLP projector in both supervised and unsupervised learning methods, and are the first to identify its key importance to model transferability to the best of our knowledge.

\section{Transferability Analysis of the Unsupervised and Supervised Pretraining Methods} \label{sec:MLP_matters}
\subsection{The Concept Generalization Task}
\label{sec-setting}

We use the concept generalization task~\cite{sariyildiz2021concept} to analyze the transferability gap between the unsupervised and supervised pretraining methods. 

\noindent \textbf{Data preparation.} Sariyildiz \etal~\cite{sariyildiz2021concept} evaluated the transferability of methods when the pretraining and evaluation dataset have semantic distance. Their experimental results show that larger semantic distance will lead to more accuracy differences among different pretraining methods.
Therefore, we enlarge the semantic gap between the pretraining and the evaluation dataset to help us compare different pretraining methods. Sariyildiz\etal~\cite{sariyildiz2021concept} use the hierarchy in WordNet~\cite{miller1998wordnet} and divide ImageNet-21K~\cite{deng2009imagenet} into six class-exclusive datasets with different semantic distance -- one for pretraining, and others for evaluation. 
Without loss of generality, we construct a smaller pretraining dataset (pre-D) and evaluation dataset (eval-D) based on ImageNet-1K~\cite{russakovsky2015imagenet} to reduce the experimental burden. Pre-D contains 652 classes mostly of organisms, and eval-D contains the other 348 classes of instrumentality.

\noindent \textbf{Transferability evaluation.} Following~\cite{sariyildiz2021concept}, to assess the transferability, we freeze all parameters in the pretrained backbone \footnote{All experiments in Sec.~\ref{sec:MLP_matters} and Sec.~\ref{MLP_enhance_SL} are conducted with ResNet50.}, and finetune the classifier with the ImageNet-1K training samples in eval-D for reporting top-1 accuracy on ImageNet-1K validation samples in eval-D.

\begin{figure}[t]
    \centering
    \includegraphics[width=\linewidth]{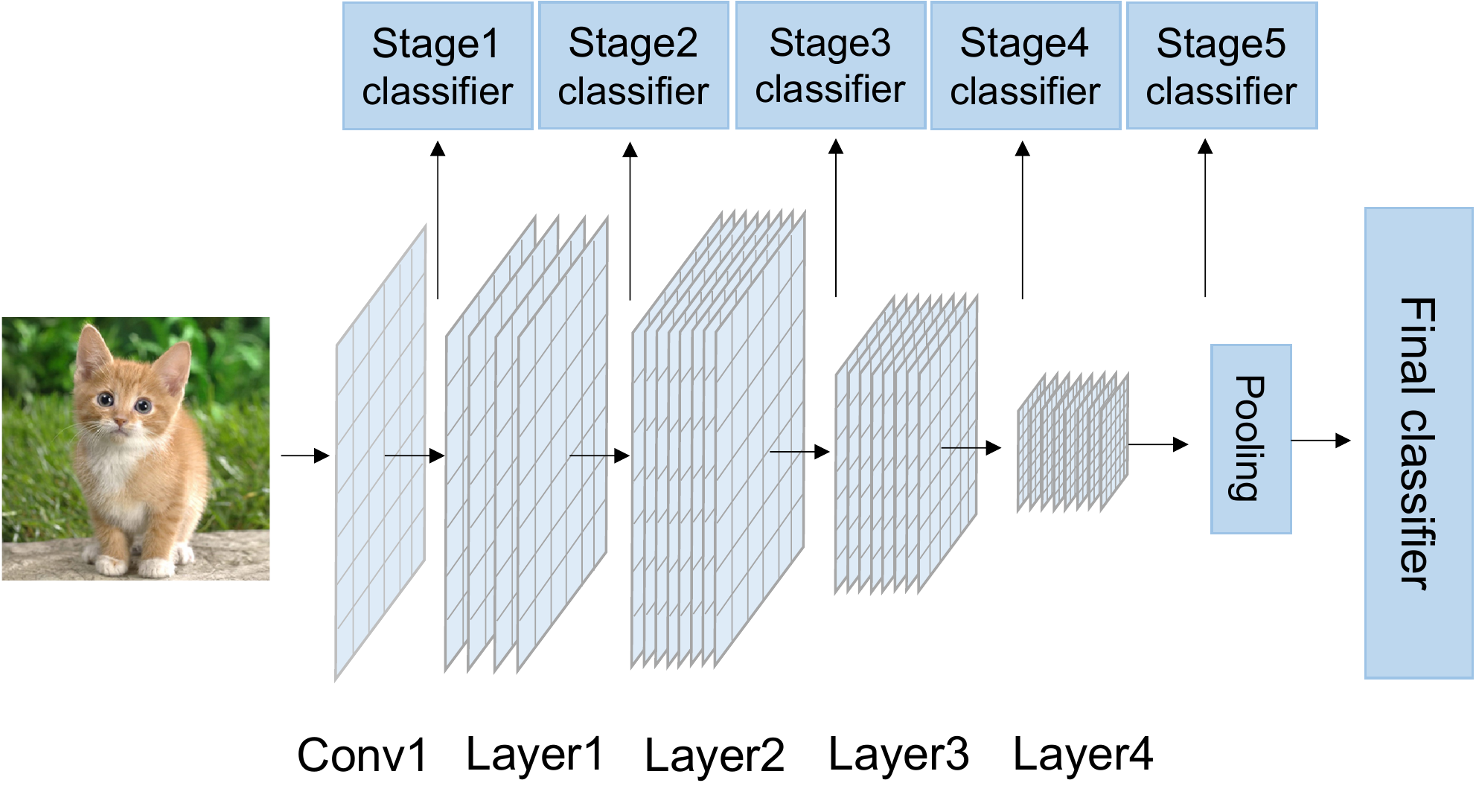}
    \vspace{-1.7em}
    \caption{\small{Schematic illustration of stage-wise evaluation. We flatten intermediate feature maps from different stages and then use them to train stage-wise classifiers. Top-1 accuracy is reported by evaluating images in eval-D with the stage-wise classifiers.}}
    \label{fig:main_stagewise_eval_illustrate}
\end{figure}

\subsection{Stage-wise Evaluation on Existing Methods}
\label{2.1} 
Motivated by works~\cite{zhao2021what,islam2021a,yosinski2014transferable} 
we make a thorough stage-wise investigation of the conventional supervised pretraining method (SL) and the existing representative unsupervised pretraining methods (Mocov1~\cite{he2020momentum}, Mocov2~\cite{chen2020improved}, Byol~\cite{grill2020bootstrap}) by evaluating the transferability of intermediate feature maps (Fig.~\ref{fig:main_stagewise_eval_illustrate}). 
After pretraining the model on pre-D, we freeze all model parameters and use the extracted intermediate feature maps of images in eval-D to finetune a stage-wise classifier for a stage-wise linear evaluation.

\begin{figure}[t]
    \centering
    \includegraphics[width=0.88\linewidth]{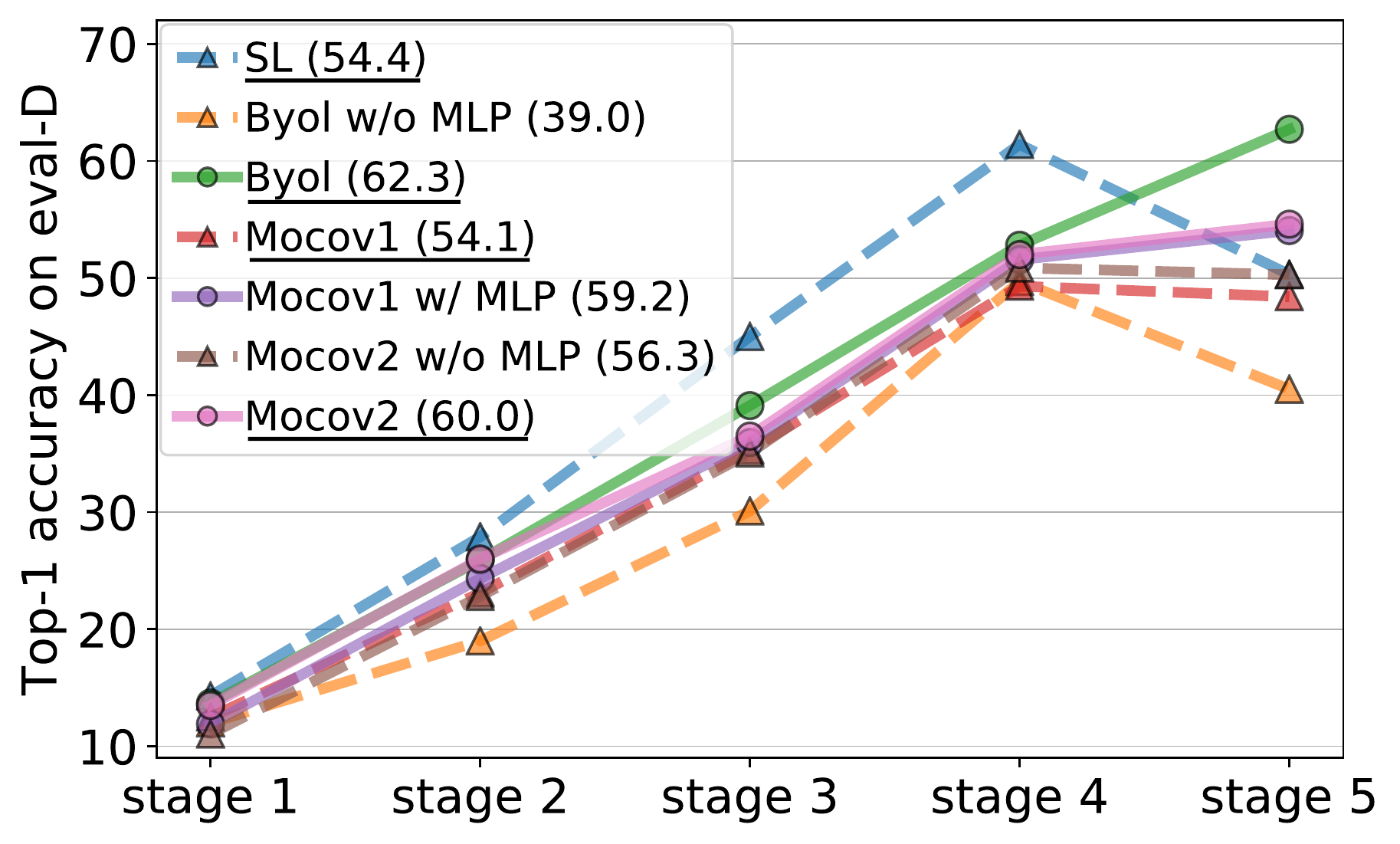}
    \caption{\small{Top-1 accuracy of stage-wise evaluation. All methods use ResNet50 as their backbones and are trained by 300 epochs with the setting in original papers. The results of linear evaluation of layer4-pooled-features (see Fig.~\ref{fig:main_stagewise_eval_illustrate}) are reported in the legend.}\vspace{-0.1em}}
    \label{fig:main_stagewise_eval_result}
\end{figure}

The evaluation results of these existing methods are depicted in Fig.~\ref{fig:main_stagewise_eval_result} (\underline{underlined} on the legend). 
Our stage-wise evaluation shows two new findings that have not been reported by existing works. First, on stage-wise evaluation from stage 1 to stage 4, SL is consistently higher than Byol, Mocov1, and Mocov2, which suggests that the semantic information in annotations can benefit the transferability of low/middle-level feature maps.  Second, on stage-wise evaluation from stage 4 to stage 5, the performance of Byol and Mocov2 still increase while SL and Mocov1 have a transferability drop. By carefully inspecting these methods, we notice an architectural difference between SL, Mocov1, Mocov2, and Byol after stage 5: An MLP projector is inserted after stage 5 in Byol and Mocov2, which does not exist in SL and Mocov1. Such difference, together with the experimental results in Fig.~\ref{fig:main_stagewise_eval_result}, leads to a new hypothesis that the MLP projector might be the core factor of the desirable transferability of unsupervised pretraining.

\subsection{MLP Improves the Transferability of Unsupervised Pretraining Methods}  
To confirm our hypothesis of the effectiveness on unsupervised pretraining methods, we ablate the MLP projectors on existing unsupervised methods, \footnote{We do not directly compare Mocov1 with Mocov2 because Mocov2 has more augmentations and the different learning rate schedule.} using stage-wise evaluation. Specifically, we remove the MLP projector in Byol and Mocov2 as Byol w/o MLP and Mocov2 w/o MLP, and add an MLP projector in Mocov1 as Mocov1 w/ MLP. The stage-wise evaluation results of these ablations are summarized in Fig.~\ref{fig:main_stagewise_eval_result}. We use solid lines for methods that have an MLP projector and dash lines for those that do not.

These ablation results offer us two observations.
First, when evaluating the layer4-pooled-features (depicted in the legend), unsupervised learning methods with an MLP projector achieve better transferability than their variants without the MLP projector, \eg, Byol, Mocov1 w/ MLP, Mocov2 achieve higher accuracy than Byol w/o MLP, Mocov1, and Mocov2 w/o MLP by $+23.3\%$, $+5.1\%$ and $+3.7\%$, respectively. Second, on stage-wise evaluation from stage 4 to stage 5, the MLP projector can help unsupervised learning methods without the MLP projector to avoid the transferability drop. 
These consistent improvements by adding an MLP projector empirically show that the MLP projector is important for the transferability of unsupervised pretraining. While there might exist some other non-linear structures that can boost the transferability, we only explore from an MLP perspective in this paper because of its simplicity and demonstrated effectiveness.

\section{MLP Can Enhance Supervised Pretraining}\label{MLP_enhance_SL}

\subsection{SL-MLP: Adding an MLP Projector to SL} \label{sec:SL-MLP method}
\begin{figure}
    \centering
    \includegraphics[width=1\linewidth]{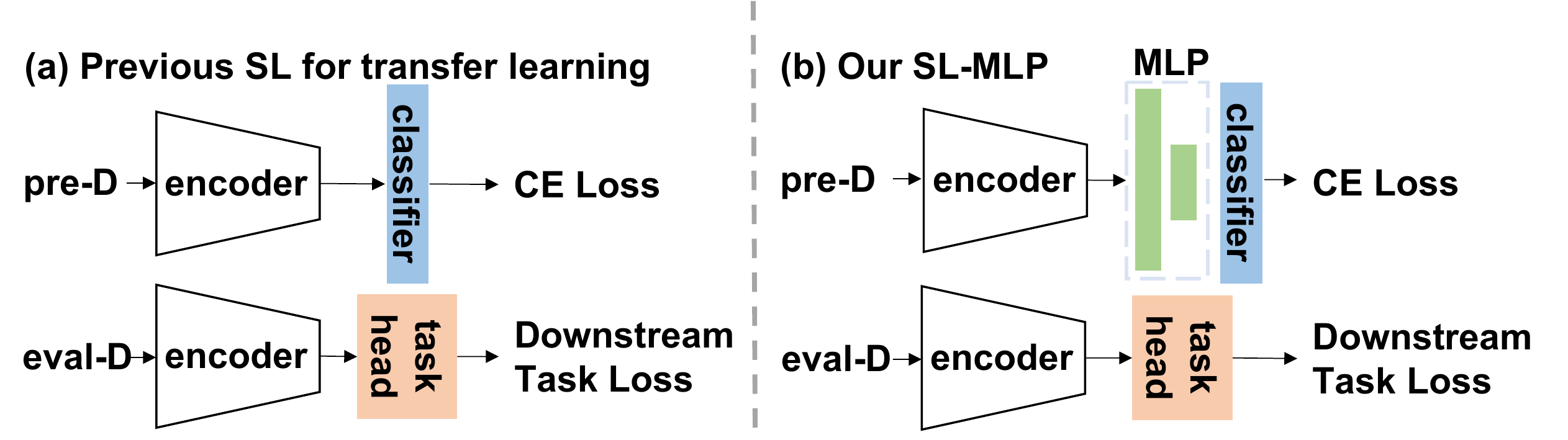}
    \vspace{-1.5em}
    \caption{\small{The difference between SL and SL-MLP. Our SL-MLP adds an MLP before the classifier compared to SL. Only the encoders in both methods are utilized for downstream tasks.}\vspace{-1em}}
    \label{fig:framework}
    
\end{figure}

Motivated by the empirical results in Sec.~\ref{sec:MLP_matters}, an interesting question is whether the MLP projector can also promote the transferability of supervised pretraining? We attempt to insert an MLP projector before the classifier on SL for better transferability. We denote this supervised pretraining method as SL-MLP (see Fig.~\ref{fig:framework} for their comparison). Specifically, SL-MLP includes a feature extractor $f(\cdot)$, an MLP projector $g(\cdot)$, and a classifier $\mathbf{W}$. Given an input image $\mathbf{x}$, the feature extractor outputs a feature $\mathbf{f}=f(\mathbf{x})$.
For example, $f(\mathbf{x})$ transforms an image $\mathbf{x}$ to a 2048 dimensional feature $\mathbf{f}$ when using the ResNet-50 backbone. The MLP projector maps $\mathbf{f}$ into a projection vector $\mathbf{g}=g(\mathbf{f})$.
Following Byol, the MLP projector consists of two fully connected layers, a batch normalization layer, and a ReLU layer, which can be mathematically formulated as $g(\mathbf{f}) = fc_2(ReLU(BN(fc_1(\mathbf{f}))))\in \mathbb{R}^{D_g}$, where $fc_1
$ and $fc_2$ are fully connected layers, the hidden feature dimension in the MLP projector is set to 4096, and $D_g$ is set to 256. 
Given the label denoted by $y$ for image $\mathbf{x}$, the objective function for SL-MLP can be formulated as
\begin{equation}
    \mathcal{L}(\mathbf{x}) = \text{CE}(\mathbf{W} \cdot g(f(\mathbf{x})), y),
\end{equation}
where $\text{CE}(\cdot)$ is the cross-entropy loss. Same as SL, only the learned feature extractor $f(\cdot)$ is utilized in downstream transfer tasks after supervised pretraining. 

\begin{figure}
    \centering
    \includegraphics[width=\linewidth]{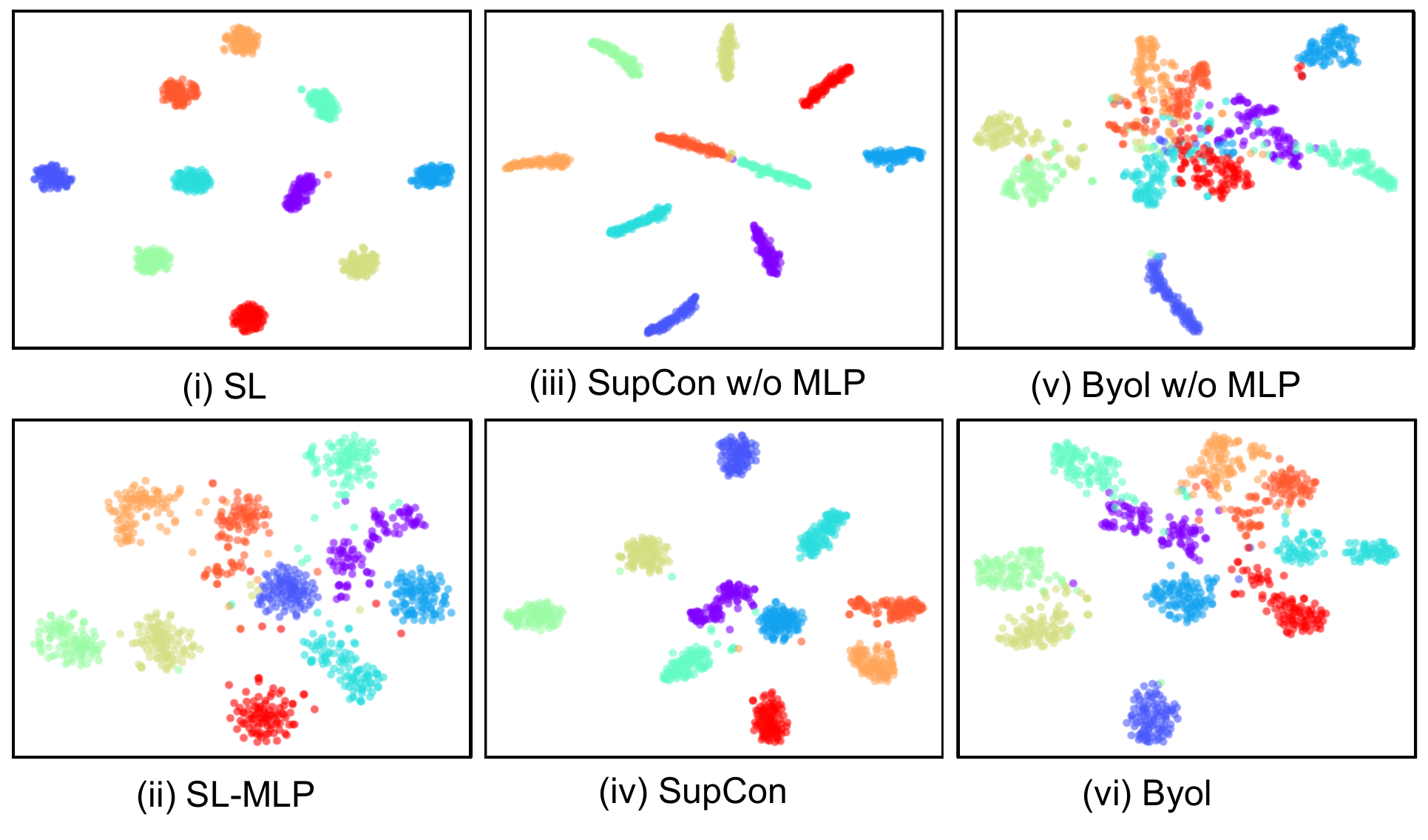}\vspace{-0.5em}
       \caption{Visualization of different methods with 10 randomly selected classes on pre-D. Different colors denote different classes. Features extracted by pretrained models without an MLP projector (top row) have less intra-class variation than those extracted by pretrained models with an MLP projector (bottom row).}
    \label{fig:intra_class_tsne}
    \vspace{-0.5em}
\end{figure}

\begin{figure}
    \centering
    \includegraphics[width=\linewidth]{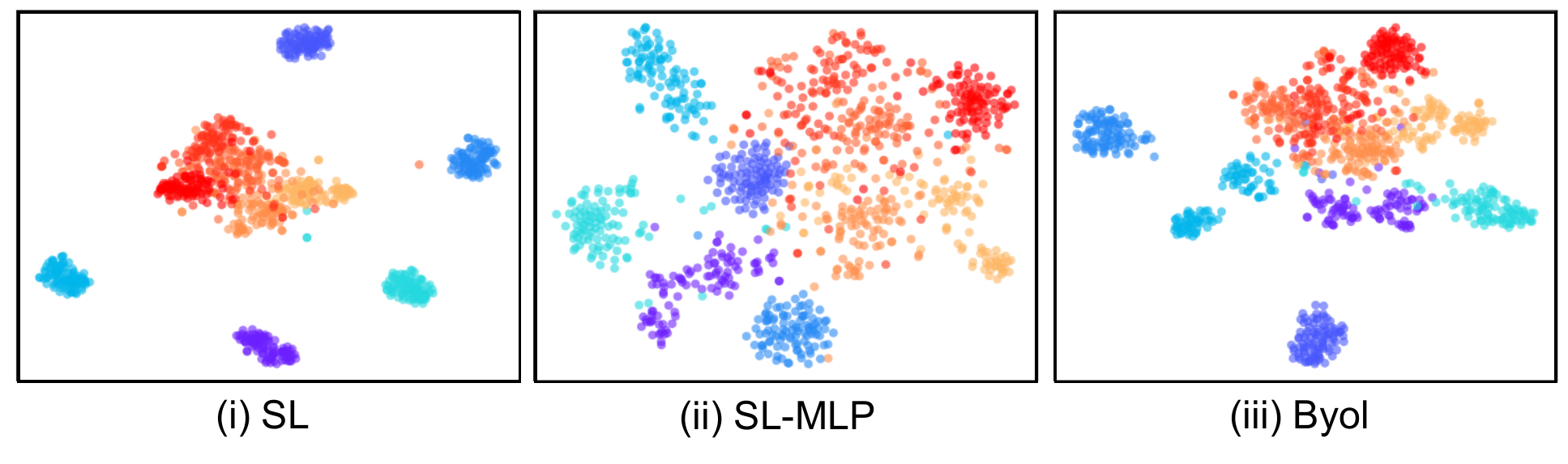}
    \vspace{-2em}
    \caption{Visualization of Feature Mixtureness between pre-D and eval-D. 
    Cold colors denote features from 5 classes that are randomly selected from pre-D, and warm colors denote features from 5 classes that are randomly selected from eval-D. \vspace{-1em}
    }
    \vspace{-0.2em}
    \label{fig:mixtureness_tsne}
\end{figure}


\begin{figure*}
    \centering
    \subfloat[]
    {\includegraphics[width=.32\linewidth]{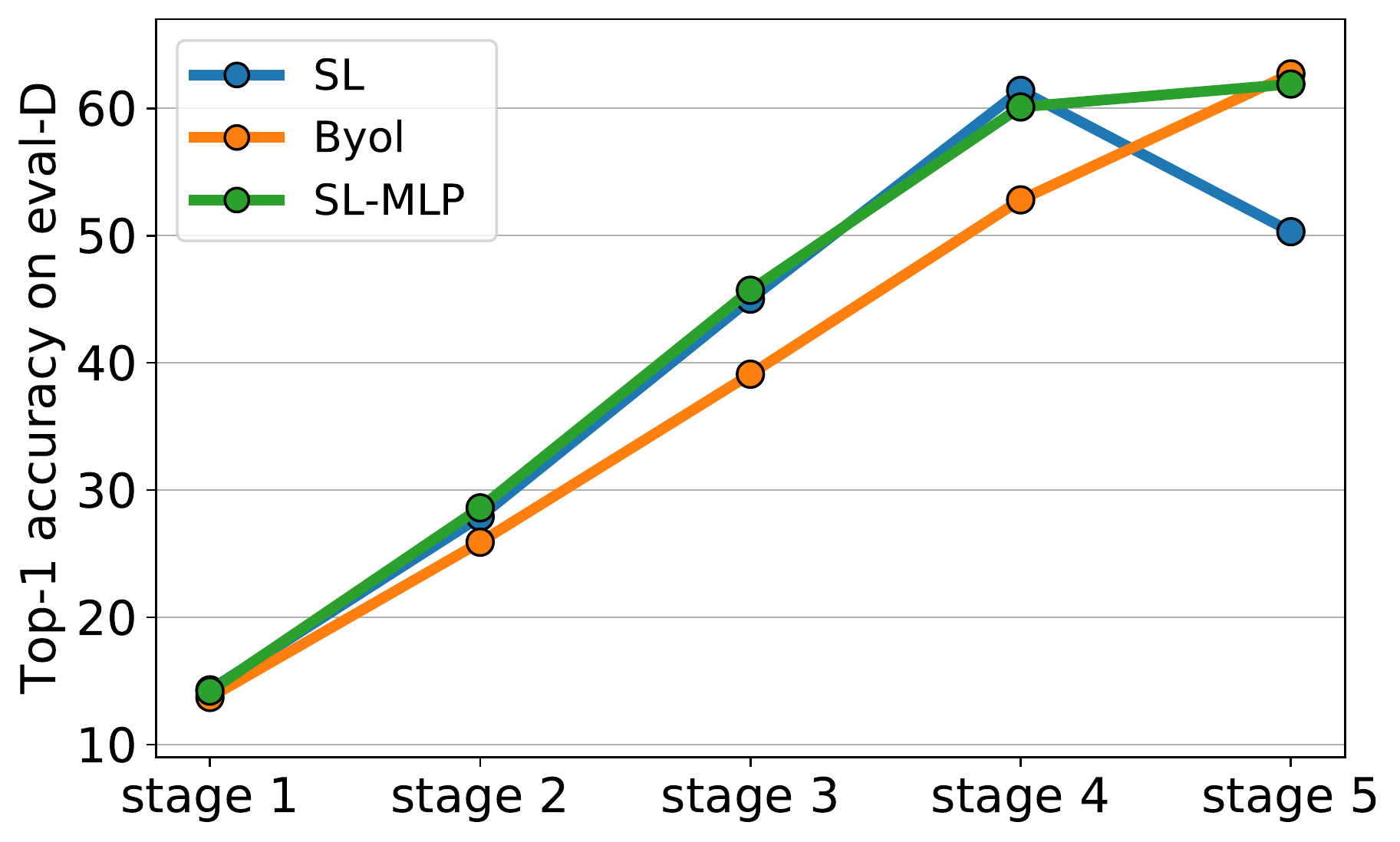}\vspace{0.3cm}}\hfill
    \subfloat[]
    {\includegraphics[width=.32\linewidth]{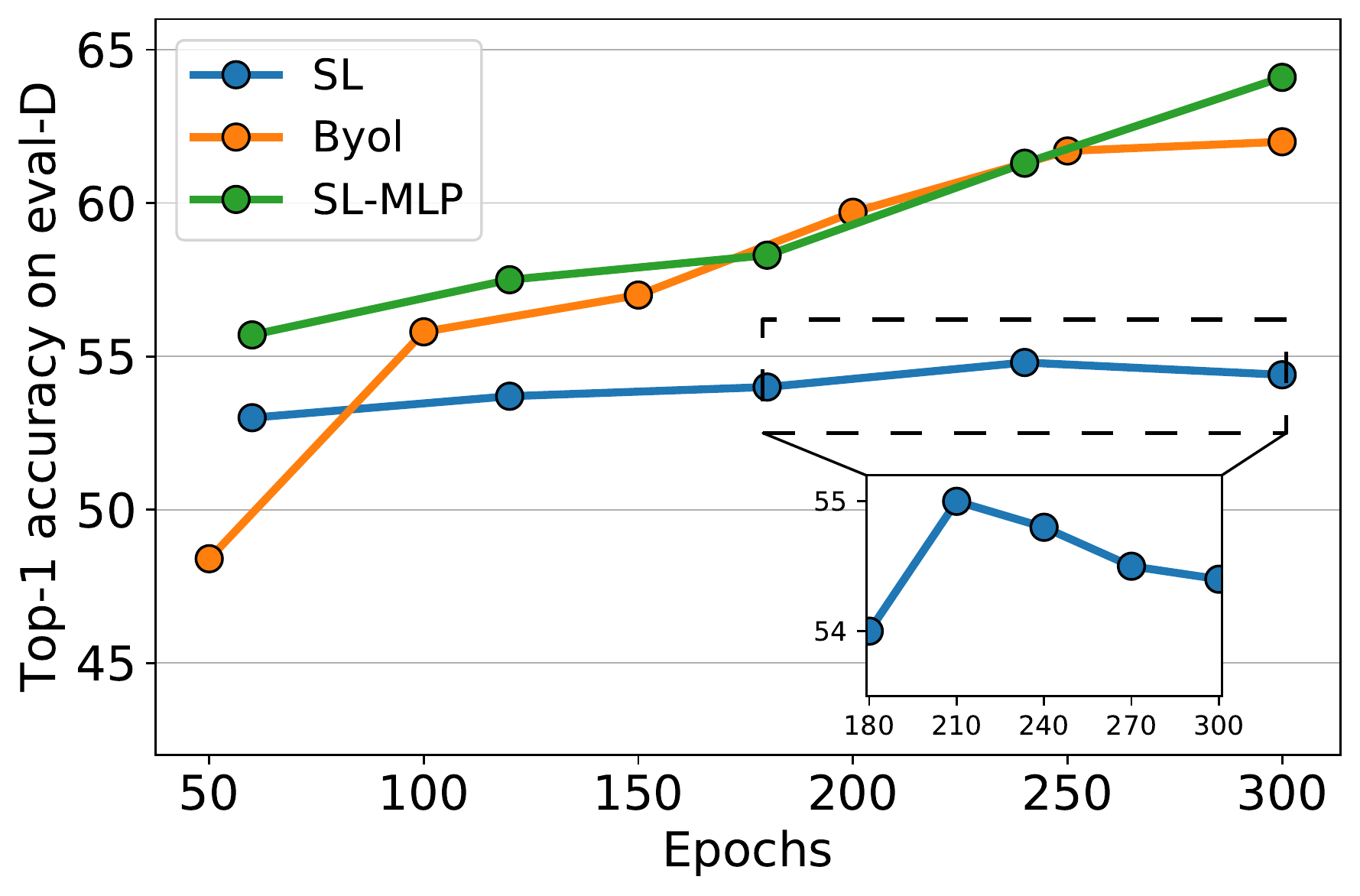}\vspace{-0.05cm}}\hfill
    \subfloat[]{\includegraphics[width = .32\linewidth]{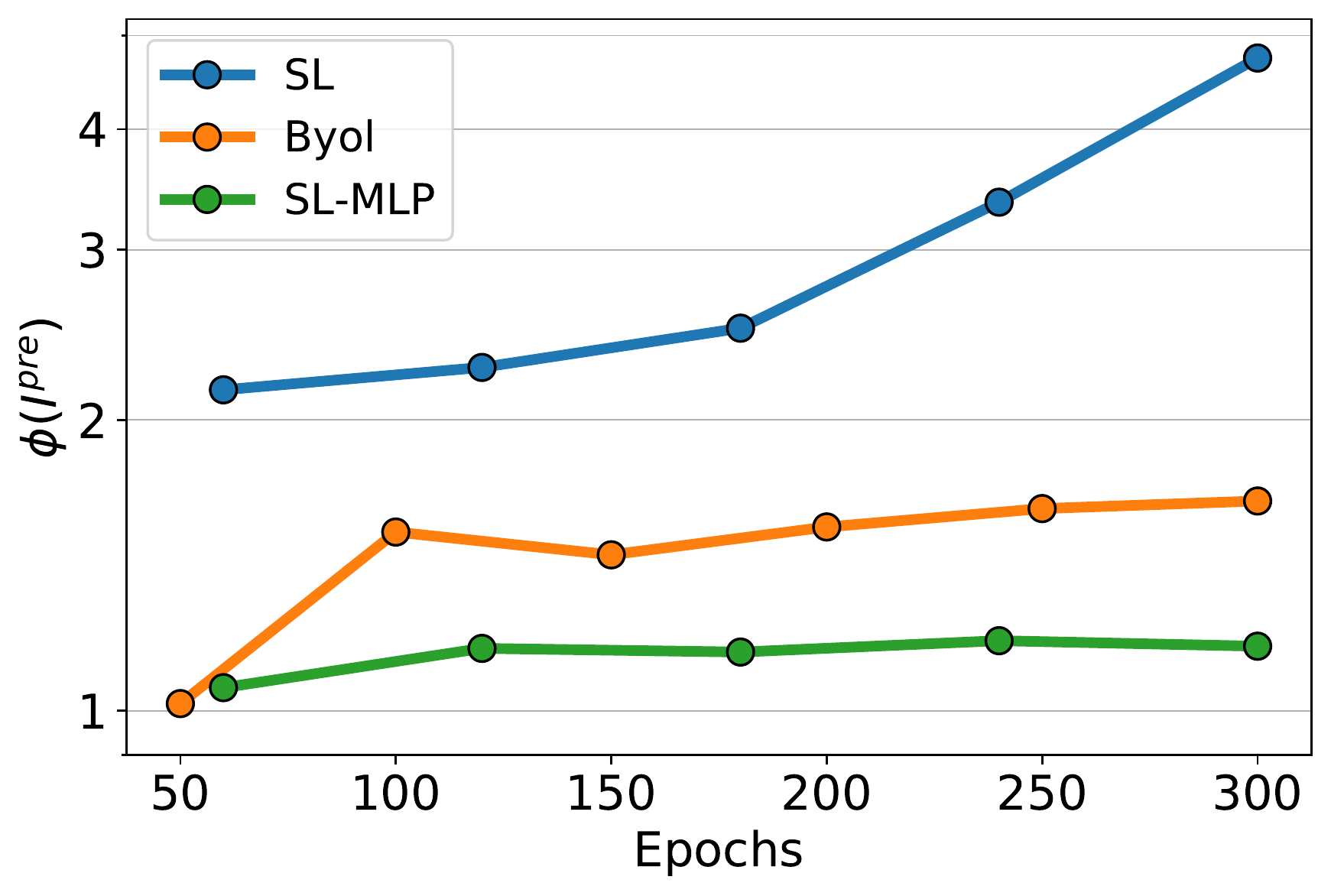}\vspace{-0.1cm}}\vspace{-0.32cm}
    \caption{\small{\textbf{(a)} Stage-wise evaluation on eval-D. \textbf{(b)} Linear evaluation accuracy on eval-D. \textbf{(c)} Discriminative ratio of features on pre-D. Following ~\cite{he2019rethinking,grill2020bootstrap}, we pretrain SL, SL-MLP, and Byol for 300 epochs.}}
    \label{fig:Stage-wise-top1-discriminative}
\end{figure*}

\begin{figure*}[ht]
\vspace{-3em}
\begin{floatrow}
\floatbox{figure}[1.37\linewidth]
{\caption{\small{\textbf{(a)} Feature Mixtureness between pre-D and eval-D. \textbf{(b)} Redundancy $\mathcal{R}$ of pretrained features during different epochs. Following ~\cite{he2019rethinking,grill2020bootstrap}, we pretrain SL, SL-MLP, and Byol for 300 epochs.}\vspace{-1em}}
 \label{fig:mixtureness-redundancy} \vspace{1cm}}
{
\subfloat[]
    {\includegraphics[width=.5\linewidth]{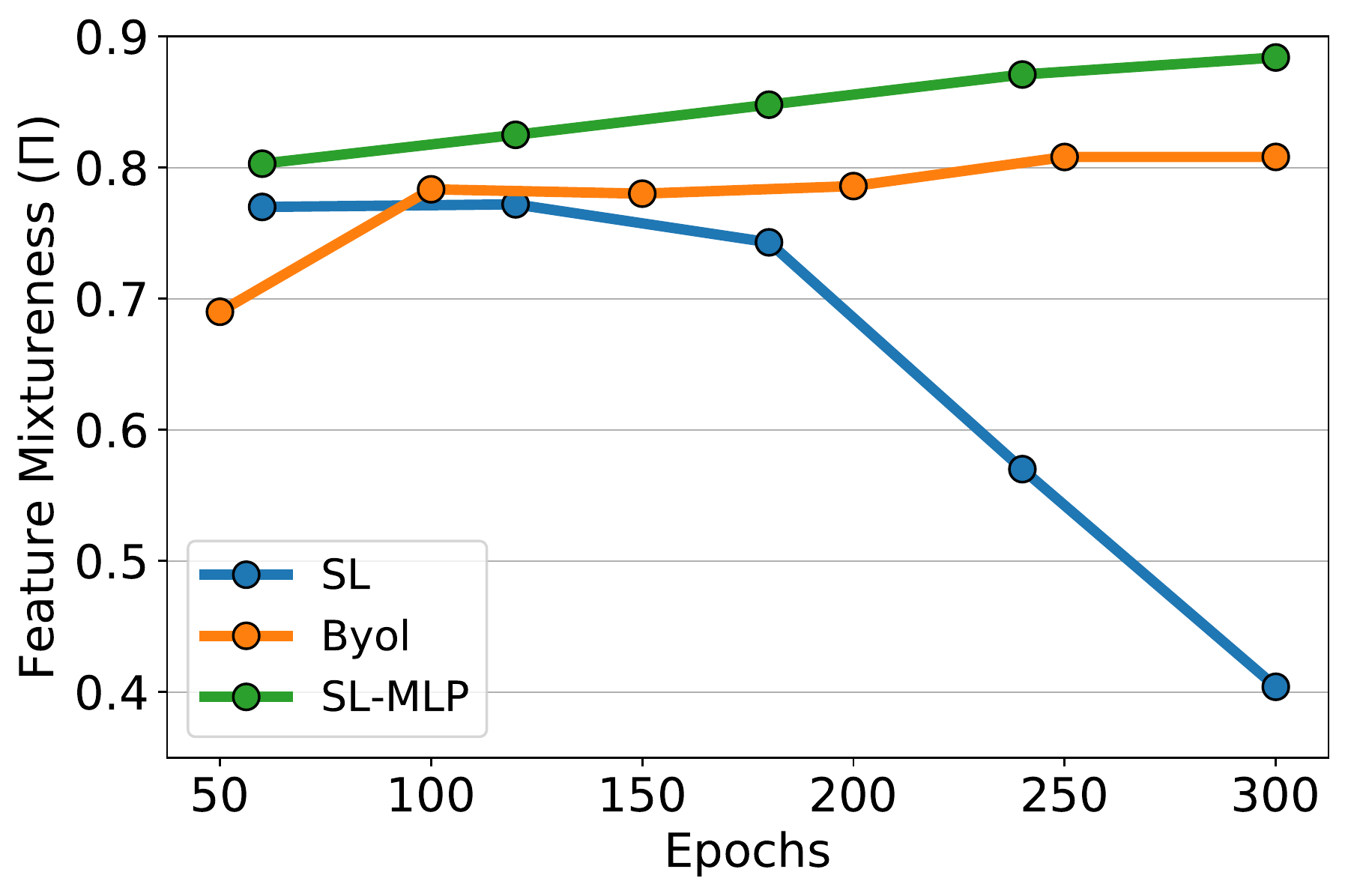}\vspace{0cm}\hspace{0cm}}\hfill
    \subfloat[]
    {\includegraphics[width=.5\linewidth]{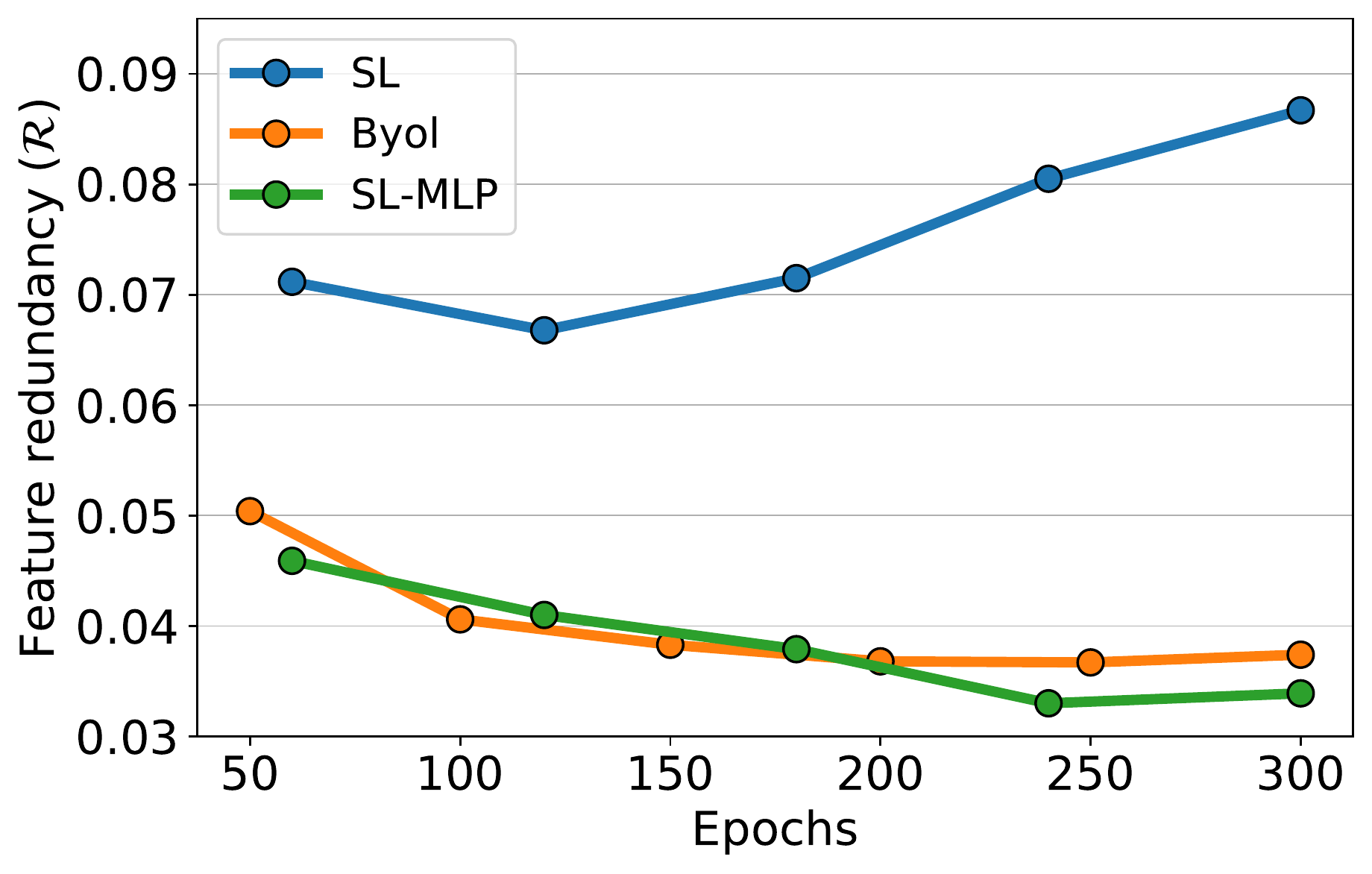}\vspace{0.05cm}}\vspace{-0.32cm}}
\hspace{-0.5cm}
\floatbox{figure}[.67\linewidth]
{\captionof{table}{\small{Redundancy $\mathcal{R}$ of pretrained features. Methods with an MLP obtain lower channel redundancy and transfer better.}\vspace{-1em}}
 \label{tab:redundancy-mlp}}
{\scriptsize
    \begin{tabular}{lccccc}
        \toprule
        Method &  EP & Top-1($\uparrow$) & $\mathcal{R}(\downarrow)$ \\
        \midrule
        SL           & 100   &55.9  & 0.078 \\
        SL-MLP      & 100   &\textbf{63.1}   & \textbf{0.035} \\
        \midrule
        SL          & 300   &54.4  & 0.087 \\
        SL-MLP      & 300   &\textbf{64.1}  & \textbf{0.034} \\
        \midrule
        Byol w$/$o MLP       & 300   &39.0  & 0.247 \\
        Byol       & 300   &\textbf{62.3}    & \textbf{0.037} \\
        \midrule
        Mocov1       & 300   &54.1 & 0.069 \\
        Mocov1 w$/$ MLP      & 300   &\textbf{59.2} & \textbf{0.058} \\
        \bottomrule
        \vspace{1em}
    \end{tabular}\vspace{-1cm}}
\end{floatrow}
\end{figure*}

\subsection{Empirical Findings of MLP in SL-MLP}
\label{sec-finding-SLMLP}
\noindent \textbf{MLP projector avoids transferability drop at stage 5 in supervised pretraining.} 
We conduct stage-wise evaluation as Sec.~\ref{2.1} again to see whether the transferability drop from stage 4 to stage 5 exists in SL-MLP. In Fig.~\ref{fig:Stage-wise-top1-discriminative}(a), the transferability of SL-MLP continuously increases from stage 1 to 5, avoiding a decrease at stage 5 as SL. 
Besides, we observe that the transferability of SL-MLP is higher than that of Byol from stage 1 to 4, indicating that annotations benefit the transferability of intermediate feature maps.

\noindent \textbf{MLP projector enlarges the intra-class variation of features.}
According to~\cite{zhao2021what,islam2021a}, features with large intra-class variation can preserve more instance discriminative information, which is beneficial for transfer learning. We reveal that adding an MLP projector also can enlarge the intra-class variation. We compare two supervised pretraining methods, \emph{i.e.,} SL, SupCon~\cite{khosla2020supervised}, and one unsupervised pretraining method, \emph{i.e.,} Byol, with their variants with/without MLP. 
Qualitatively, we visualize their features learned on pre-D by t-SNE in Fig.~\ref{fig:intra_class_tsne}. The intra-class variation of features from SL-MLP, SupCon, and Byol are larger than that from SL, SupCon w/o MLP, and Byol w/o MLP, respectively. 
Quantitatively, following LDA~\cite{balakrishnama1998linear}, we utilize a discriminative ratio $\phi(I^{pre})$ to measure intra-class variation on pre-D, where $I^{pre}$ denotes pre-D (mathematically defined in Sec.~\ref{sec:theoretical_analysis_sl_mlp}). Smaller discriminative ratio $\phi$ usually means larger intra-class variation\footnote{Strictly speaking, larger intra-class variation is relative to inter-class distance, which is theoretically defined as discriminative ratio. We use ``intra-class variation'' to be consistent with previous work~\cite{islam2021a,zhao2021what}.}. Comparing Fig.~\ref{fig:Stage-wise-top1-discriminative}(c) with Fig.~\ref{fig:Stage-wise-top1-discriminative}(b), we can see Byol and SL-MLP have smaller $\phi(I^{pre})$ but higher accuracy on eval-D than SL, which shows larger intra-class variation can benefit transferability. Furthermore, when inspecting SL only, we can see a process where the accuracy on eval-D first rises and then descends (after 210 epochs) along with $\phi(I^{pre})$ increasing. This phenomenon can be theoretically explained in Sec.~\ref{sec:theoretical_analysis_sl_mlp}. We additionally provide the visualization of intra-class variation on different pretraining epochs in Supplementary B.1.

\noindent \textbf{MLP projector reduces feature distribution distance between pre-D and eval-D.} According to~\cite{blitzer2008learning,liu2019towards}, decreasing the feature distribution distance between pre-D and eval-D in the representation space can benefit transfer learning. Intuitively, the distribution distance between two sets of features is small when features are well mixed (visualization provided in Supplementary A). Therefore, we compare the mixtureness of features in pre-D and eval-D to indicate the feature distribution distance between SL and SL-MLP. Graphically, we visualize features from pre-D and eval-D by t-SNE in Fig.~\ref{fig:mixtureness_tsne}. We observe that features from pre-D and eval-D are more mixed comparing SL and SL-MLP, indicating that MLP projector can mitigate the distribution distance between pre-D and eval-D. 
Quantitatively, we define \textit{Feature Mixtureness} $\Pi$ in the feature space as
\begin{equation}
    \scriptsize
    \Pi = 1-\frac{1}{C}\sum_{i=1}^C \left |\frac{top_k^{eval}(i)}{k} - \frac{C^{eval}}{C} \right|,
\end{equation}
where $C=1000$ is total number of classes in ImageNet-1K, $C^{eval}$ represents the number of classes in eval-D, and $top^{eval}_k(i)$ represents the number of classes in eval-D found by top $k$ neighbors search of any class $i \in C$. Since the percentage of finding a sample from eval-D in $k$ nearest neighbors is $C^{eval}/C$ when pre-D and eval-D are uniformly mixed, Feature Mixtureness measures the similarity of the current and the uniformly mixed distribution between pre-D and eval-D in the feature space. We examine Feature Mixtureness of SL, SL-MLP, and Byol during different pretraining epochs in Fig.~\ref{fig:mixtureness-redundancy}(a). Feature Mixtureness of SL gradually decreases, which indicates that SL will enlarge the distribution difference between pre-D and eval-D. In contrast, SL-MLP and Byol show consistently high Feature Mixtureness, indicating that the MLP projector can reduce the distribution distance between pre-D and eval-D.
We visualize the evolution of Feature Mixtureness in Supplementary B.2.

\noindent \textbf{MLP projector reduces feature redundancy.}
Inspired by~\cite{zbontar2021barlow}, high channel redundancy limits the capability of feature expression in deep learning. 
Mathematically, we compute Pearson correlation coefficient among feature channels to evaluate \textit{feature redundancy} $\mathcal{R}$, \emph{i.e,}
\begin{equation}
    \scriptsize
    \mathcal{R} = \displaystyle \frac{1}{d^2}\sum_{i=1}^{d}\sum_{j=1}^{d}|\rho(i,j)|, \quad \rho(i,j)\!=\!\frac{\sum_{n=1}^N\mathbf{f}_{n,i} \cdot \mathbf{f}_{n,j}}{\sqrt{\sum_{n=1}^N||\mathbf{f}_{n,i}||^2}\sqrt{\sum_{n=1}^N||\mathbf{f}_{n,j}||^2}}
\end{equation}
where $d=2048$ is the feature dimension, $\rho(i,j)$ is Pearson correlation coefficient of feature channel $i$ and $j$. As shown in Fig.~\ref{fig:mixtureness-redundancy}(b), SL-MLP has lower feature redundancy than SL, which indicates that the MLP projector can reduce feature redundancy. In Tab.~\ref{tab:redundancy-mlp}, we further confirm that the MLP projector can reduce the feature redundancy and thus increase the accuracy on eval-D by ablating the MLP projector on various pretraining methods.

\begin{figure}
    \centering
    \includegraphics[width=0.9\linewidth]{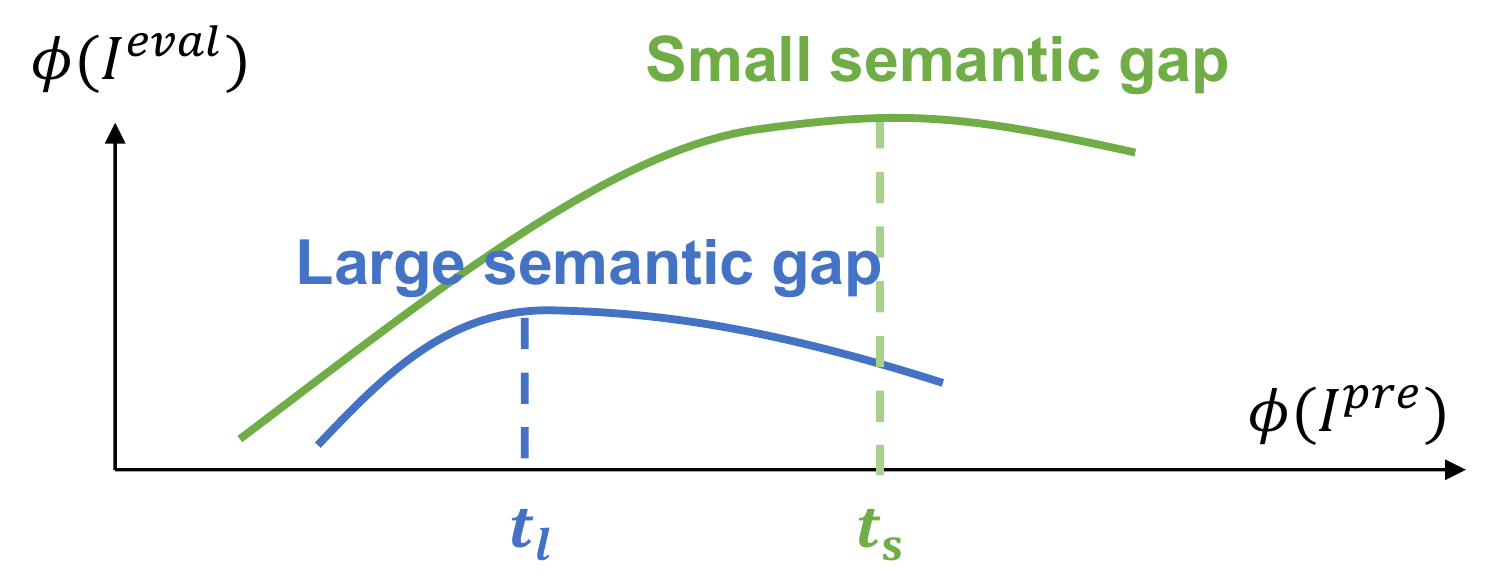}
    \caption{\small{Insights for transferability. $\phi(I^{pre})$ and $\phi(I^{eval})$ are the discriminative ratios (Eq. \ref{eq:discriminative_ratio}) on the pretraining and evalution datasets. Higher $\phi(I^{eval})$ indicates better model transferability. Green and Blue line show the performance curve on the evaluation dataset with small and large semantic gap, respectively.}\vspace{-0.5em}}
    \label{fig:insight_illustration}
\end{figure}

\subsection{Theoretical Analysis for Empirical Findings} \label{sec:theoretical_analysis_sl_mlp}

In this section, we provide a theoretical analysis to reveal that: 1) maximizing the discriminative ratio $\phi(I^{pre})$ of a model on the pretraining dataset above a certain threshold will lead to a transferability descrease (shown by blue/green lines in Fig.~\ref{fig:insight_illustration}); 2) the threshold is smaller when the semantic gap between the pretraining and evalutaion dataset is larger ($t_l<t_s$ in Fig.~\ref{fig:insight_illustration}).



Mathematically, the discriminative ratio $\phi(I)$ on the dataset $I$ can be defined by LDA~\cite{balakrishnama1998linear} as
\begin{equation} \label{eq:discriminative_ratio}
    \small
    \phi(I)=D_{inter}(I)/D_{intra}(I),
\end{equation}
where $D_{inter}(I)\!=\!\frac{1}{C(C-1)}\sum_{j=1}^{C}\sum_{k=1,k\neq j}^C  || \mu(I_j)\!-\!\mu(I_k)||^2$ is the inter-class distance, $D_{intra}(I)=\frac{1}{C}\sum_{j=1}^{C}(\frac{1}{|I_j|}\sum_{(x_i,y_i)\in I_j}  || \mathbf{f}_i-\mu(I_j)||^2)$ is the intra-class distance, and $C$ is the number of classes. $\mu(I_j)=\frac{1}{I_j}\sum_{(x_i,y_i)\in I_j} \mathbf{f}_i$ is the center of features in class $I_j$, and $\mathbf{f}$ is the feature in Sec.~\ref{sec:SL-MLP method}.
Higher discriminative ratio $\phi$ indicates higher classification accuracy. Inspired by~\cite{liu2020negative}, we analyze the relation between $\phi(I^{pre})$ and $\phi(I^{eval})$ in Theorem 1 (Supplementary C).
\begin{theorem}
Given $\phi_1(I^{pre})<\phi_2(I^{pre})$, $\phi_1(I^{eval})>\phi_2(I^{eval})$ when $\phi_1(I^{pre})>t$, where $t$ is a threshold that is negatively related to the feature distribution distance.
\end{theorem}
\noindent \ul{\emph{Insights for the intra-class variation.}} 
Theorem 1 reveals that continuously minimizing the intra-class variation (maximizing the discriminative ratio) on the pretraining dataset will decrease the transferability of the model when 
the discriminative ratio $\phi(I^{pre})$ is larger than $t$. It explains the observation in Fig.~\ref{fig:Stage-wise-top1-discriminative}(b) and Fig.~\ref{fig:Stage-wise-top1-discriminative}(c) that training with more than 210 epochs leads to better performance on pre-D, but a worse transferability on eval-D. This insight suggests that we should not make the intra-class variation on the pretraining dataset too small when designing the objective function or network architecture (\emph{e.g.,} adding an MLP projector).

\noindent \ul{\emph{Insights for the relation between the feature distribution distance and threshold $t$.}} 
When the feature distribution distance between the pretraining and evaluation dataset is large, the threshold $t$ is small, in which case it is easier to have the undesirable effect of increasing the discriminative ratio $\phi(I^{pre})$ on pre-D leading to decreasing the discriminative ratio $\phi(I^{eval})$ on eval-D (and thus the accuracy on the evaluation data). This insight suggests that we should maintain more intra-class variation on the pretraining dataset when transferring the model to a target dataset which has a larger semantic distance from the pretraining dataset.   

\section{Experiment\label{experiments}}
\subsection{Experimental Setup}

\noindent \textbf{Datasets}. For backbone analysis, we keep using the concept generalization setting described in Sec. \ref{sec-setting}. For generalization to other classification tasks, we follow the setup in ~\cite{islam2021a}, which pretrains the models on the whole ImageNet-1K dataset and then evaluates the transferability on 12 classification datasets~\cite{wang2017chestx,mohanty2016using,olsen2019deepweeds,cimpoi2014describing,helber2019eurosat,tian2020kaokore,lake2015human,cheng2017remote,wang2019learning,netzer2011Reading,nilsback2008automated,codella2019skin} from different domains.
Furthermore, the COCO~\cite{lin2014microsoft} dataset is used to evaluate the performance of SL-MLP pretrained by ImageNet-1K~\cite{russakovsky2015imagenet} on object detection task.

\noindent \textbf{Details}. 
For SL and SL-MLP pretraining, the cross-entropy is deployed as the loss function. The MLP projector deployed in SL-MLP is described in Sec.~\ref{sec:SL-MLP method}.
Following \cite{he2016deep}, we use the SGD optimizer with a cosine decay learning rate of 0.4 to optimize SL and SL-MLP, and set the batch size to 1024. 
Byol is used as a representative method for comparisons in backbone analysis and object detection. Following~\cite{grill2020bootstrap}, we use LARS optimizer~\cite{you2017scaling} with a cosine decay learning rate schedule and a warm-up of 10 epochs to pretrain the network. The initial learning rate is set to 4.8. We set the batch size to 4096 and the initial exponential moving average parameter $\tau$ to 0.99. Except for the backbone analysis, we use ResNet50 as the default backbone. More detailed pretraining setups of different backbones and different methods are provided in Supplementary H.1.

\begin{table}[t]
  \centering
  \footnotesize
  \caption{\small{Concept generalization task. We report Top-1 accuracy on eval-D of SL-MLP, Byol, and SL on various backbones. SL-MLP and Byol share the same MLP projector. 
  }}
  \resizebox{\textwidth}{!}{
\begin{tabular}{lccccc}
    \toprule
    Method & Architecture & Labels
& MLP & Epochs & Top-1($\uparrow$)  \\
    \midrule
    SL    & ResNet50 &\checkmark & & 100   & 55.9  \\
    SL-MLP & ResNet50 &\checkmark &\checkmark & 100   & 63.1       \\
    Byol  & ResNet50 & &\checkmark & 300   & 62.3     \\
    SL    & ResNet50 &\checkmark &  & 300   & 54.4   \\
    SL-MLP & ResNet50 &\checkmark &\checkmark & 300   & \textbf{64.1}       \\
    \midrule
    SL    & ResNet34 &\checkmark & & 100  & 50.1   \\
    SL-MLP & ResNet34 &\checkmark &\checkmark  & 100   & 55.0   \\
    Byol  & ResNet34 & &\checkmark & 300   &  54.8  \\
    SL    & ResNet34 &\checkmark & & 300  & 50.2   \\
    SL-MLP & ResNet34 &\checkmark &\checkmark  & 300   & \textbf{55.8}   \\
    \midrule
    SL    & ResNet101 &\checkmark & & 100 & 56.0   \\
    SL-MLP & ResNet101 &\checkmark &\checkmark & 100   & 63.6   \\
    SL    & ResNet101 &\checkmark & & 300 & 53.9   \\
    SL-MLP & ResNet101 &\checkmark &\checkmark & 300   & \textbf{64.7}   \\
    \midrule
    SL    & Swin-tiny   &\checkmark & & 100 & 58.9   \\
    SL-MLP & Swin-tiny     &\checkmark &\checkmark   & 100   & \textbf{60.6}   \\
    \midrule
    SL    &  EfficientNetb2 &\checkmark & & 100 & 57.6   \\
    SL-MLP & EfficientNetb2 &\checkmark &\checkmark & 100   & \textbf{64.2}  \\
    \bottomrule
    \end{tabular}}%
  \label{tab:unseen_class_generalization_tasks}%
\end{table}%

\begin{table}[htbp]
  \centering
  \footnotesize 
  \caption{\small Object detection results. All methods are pretrained on ImagNet-1K, then finetuned on COCO using Mask-RCNN (R50-FPN) based on Detectron2~\cite{wu2019detectron2}. Sup. and Unsup. are short for supervised learning and unsupervised learning, respectively. Results of methods\dag~are from~\cite{xie2021detco}.}
  \small
  \resizebox{\textwidth}{!}{
    \begin{tabular}{lcccccc}
    \toprule
    Method & Sup.&Unsup.&Epoch&AP&AP50&AP75\\
    \midrule
    SL    & \checkmark      &       & 100   & 38.9  & 59.6  & 42.7 \\
    SL-MLP &  \checkmark     &       & 100   & 39.7  & 60.4  & 43.1 \\
  InsDis\dag~\cite{wu2018unsupervised} &       &    \checkmark   & 200   & 37.4  & 57.6  & 40.6 \\
   PIRL\dag~\cite{misra2020self}  &       & \checkmark      & 200   & 37.5  & 57.6  & 41.0 \\
  SwAV\dag~\cite{caron2020unsupervised}  &       & \checkmark      & 200   & 38.5  & 60.4  & 41.4 \\
  Mocov2\dag~\cite{chen2020improved} &       & \checkmark      & 200   & 38.9  & 59.4  & 42.4 \\
    Byol~\cite{grill2020bootstrap}  &       & \checkmark      & 300   & 39.4  & 60.4  & 43.2 \\
    SL & \checkmark & &300 & 40.1 & 61.1 & 43.8 \\
    SL-MLP & \checkmark      &       & 300   & \textbf{40.7}  & \textbf{61.8}  & \textbf{44.2} \\
    \bottomrule
    \end{tabular}}%
    \vspace{-1.em}
  \label{tab:detection}%
\end{table}%

\begin{table*}[t]
  \centering
  \caption{\small Linear evaluation on fixed backbone, full network finetuning, and few-shot learning performance on 12 classification datasets in terms of top-1 accuracy. All models are pretrained for 300 epochs with the same code base except for  SelfSupCon\dag~(Mocov2) which pretrained for 400 epochs using the results illustrated in~\cite{islam2021a}. Average results style: \textbf{best}, \underline{second best}.\vspace{-0.25em}}
  \resizebox{0.98\textwidth}{!}{
    \begin{tabular}{lccccccccccccc}
    \toprule
    Method & ChestX & CropDisease & DeepWeeds & DTD  & EuroSAT & Flowers102 & Kaokore & Omniglot & Resisc45 & Sketch & SVHN  & ISIC & Average \\
    \midrule
    \emph{linear evaluation} \\
    \midrule
    SL             & 45.45 & 96.80 & 84.02 & 66.22 & 95.07 & 83.69 & 75.40 & 64.14 & 85.36 & 67.82 & 67.13 & 79.58 & 75.89 \\
    SL-MLP         & 49.89 & 99.02 & 87.86 & 72.61 & 96.63 & 93.46 & 81.12 & 76.73 & 91.66 & 74.51 & 75.16 & 81.53 & \textbf{81.68} \\
    SupCon w/o MLP & 41.38 & 91.52 & 73.16 & 62.93 & 89.84 & 73.23 & 66.38 & 44.54 & 76.55 & 55.21 & 61.45 & 68.54 & 67.06 \\
    SupCon         & 47.71 & 98.79 & 85.66 & 74.20 & 95.83 & 92.24 & 79.42 & 73.42 & 91.14 & 76.80 & 74.26 & 79.78 & \underline{80.77} \\
      SelfSupCon\dag         & 48.08 & 99.06 & 87.88 & 72.71 & 96.97 & 89.62 & 81.67 & 69.66 & 90.88 & 69.12 & 69.95 & 81.51 & 79.70  \\
    \midrule
    \emph{finetuned with 1000 training samples} \\
    \midrule
    SL              & 40.86 & 94.31 & 86.95 & 62.12 & 94.05 & 88.94 & 78.22 & 46.16 & 80.32 & 14.17 & 82.16 & 78.28 & 70.54 \\
    SL-MLP          & 42.34 & 94.48 & 89.64 & 63.90 & 95.30 & 90.20 & 77.98 & 46.66 & 83.13 & 17.32 & 80.19 & 78.82 & \textbf{71.66} \\
    SupCon w/o MLP  & 41.72 & 93.52 & 84.95 & 58.09 & 95.15 & 88.23 & 78.95 & 45.68 & 80.63 & 14.39 & 82.25 & 77.96 & 70.12 \\
    SupCon          & 41.84 & 93.46 & 88.70 & 61.81 & 94.54 & 91.28 & 78.35 & 46.02 & 81.62 & 15.84 & 81.85 & 78.51 & \underline{71.15} \\ 
    SelfSupCon\dag & 43.09 & 93.95 & 88.10 & 62.95 & 95.47 & 88.92 & 79.41 & 45.33 & 81.14 & 10.57 & 82.37 & 78.27 & 70.88 \\
    \midrule
    \emph{5-ways 5-shots few-shot classification} \\
    \midrule
    SL             & 25.64 & 89.07 & 54.32 & 78.58 & 82.96 & 93.14 & 46.14 & 92.82 & 84.17 & 87.06 & 38.03 & 41.22 & 67.76\\
    SL-MLP         & 26.89 & 93.45 & 59.08 & 83.04 & 87.16 & 96.88 & 50.77 & 95.73 & 89.00 & 89.84 & 41.96 & 46.76 & \textbf{71.71}\\
    SupCon w/o MLP & 23.62 & 75.64 & 49.34 & 73.04 & 73.90 & 82.16 & 38.10 & 67.87 & 75.18 & 81.01 & 34.92 & 35.16 & 59.16\\
    SupCon         & 26.18 & 94.09 & 59.36 & 85.02 & 87.97 & 96.55 & 51.02 & 94.49 & 89.01 & 89.75 & 41.67 & 43.48 & \underline{71.55}\\
    \bottomrule
    \end{tabular}}%
    \vspace{-0.5em}
  \label{tab:12tasks}%
\end{table*}%

\begin{table}[t]
  \centering
  \footnotesize
  \caption{\small Empirical analysis of architectural design of the MLP projector. We pretrain models over 100 epochs and set the output dimension to 2048. Top-1 accuracy on eval-D is reported. }
  \resizebox{\textwidth}{!}{
    \begin{tabular}{cccccccc}
    \toprule
    \multicolumn{1}{c}{\multirow{2}[4]{*}{Exp}} & \multicolumn{4}{c}{Components} & \multicolumn{1}{c}{\multirow{2}[4]{*}{+Params}} & \multicolumn{1}{c}{\multirow{2}[4]{*}{Top-1}} 
    \\
\cmidrule{2-5}          & Input FC & BN    & ReLU  & Output FC &       &         \\
    \midrule
    (a) &       &       &       &       & /     & 55.9    \\
    (b)     & \checkmark     &       &       &       & 4.196M & 56.6   \\
    (c)     & \checkmark     & \checkmark     &       & \checkmark     & 8.395M & 61.0   \\
    (d)     & \checkmark     &       & \checkmark     & \checkmark     & 8.391M & 60.1     \\
    (e)     &       & \checkmark     & \checkmark     &       & 0.004M & 60.5   \\
    \midrule
    SL-MLP     & \checkmark     & \checkmark     & \checkmark     & \checkmark     & 8.395M & 62.5  \\
    \bottomrule
    \end{tabular}}%
    \vspace{-1.75em}
  \label{tab:structural ablation}%
\end{table}%

\subsection{Experimental Results} \label{sec:experimental_results}
\noindent \textbf{Generalize to unseen concepts with diverse backbones.}
We verify the effectiveness of the added MLP projector on SL using \emph{concept generalization task} with different backbones.
Following evaluation method mentioned in Sec.~\ref{sec-setting}, we train a linear classifier with the frozen backbone for 100 epochs, and report the top-1 accuracy on eval-D in Tab.~\ref{tab:unseen_class_generalization_tasks}. 
Firstly, SL-MLP obtains better performance than SL among different backbones. 
Specifically, with ResNet50, SL-MLP improves SL to 63.1 (+7.2\%) when we pretrain the model by only 100 epochs, which bridges the performance gap between SL and Byol. In 300 epochs setting, SL has a transferability drop compared to 100 epochs setting (55.9\%$\rightarrow$54.4\%), but the transferability of SL-MLP continue to increase (63.1\%$\rightarrow$64.1\%). 
Secondly, SL-MLP (64.1\%) performs better than Byol (62.3\%) when both are trained by 300 epochs. Experimental results in Tab.~\ref{tab:unseen_class_generalization_tasks} also confirm that SL-MLP can consistently improve the transferability of SL on various backbones, \eg ResNet101~\cite{he2016deep}, Swin-tiny\cite{liu2021swin}, and EfficientNetb2~\cite{tan2019efficientnet}. 
Swin-tiny achieves relatively smaller gain (\textbf{+1.7\%}) due to its 
good Feature Mixtureness (0.86), which is close to SL-MLP
in Fig.~\ref{fig:mixtureness-redundancy}(a).

\noindent \textbf{Generalize to other classification tasks.}
To evaluate if the added MLP can help SL to transfer better on cross-domain tasks, following~\cite{islam2021a}, we pretrain the model on ImageNet-1K, and evaluate the transferability on 12 classification datasets from different domains. As illustrated in Tab.~\ref{tab:12tasks}, supervised pretraining methods with the MLP projector, \emph{i.e.,} SL-MLP and SupCon~\cite{khosla2020supervised}, outperform their no MLP counterparts, \emph{i.e.,} SL and SupCon w/o MLP on linear evaluation, by 5.79\%, 13.71\% on the averaged Top-1 accuracy, respectively. Consistent results can be observed on finetuning and few-shot learning settings.
More results are provided in Supplementary I. 
Besides, by comparing SupCon, SL-MLP and SupCon w/o MLP, SL, we conclude that the MLP projector instead of the contrastive loss plays the key role in increasing transferability. Our conclusion contrasts with previous works~\cite{zhao2021what,islam2021a} because they ignore the MLP projector before the contrastive loss.

\noindent \textbf{Generalize to object detection.}
We assess the transferability improvement by the MLP projector on COCO object detection task. We follow the settings in~\cite{he2020momentum} to finetune the whole network with $1\times$ schedule.
In Tab.~\ref{tab:detection}, we report results using Mask-RCNN (R50-FPN), as detailed in Supplementary H.
When both are pretrained over 100 epochs, SL-MLP performs better than SL (without MLP) on object detection by \textbf{+0.8 AP}. If MLP is used by both supervised and unsupervised pretraining, SL-MLP pretrained by 100 epochs achieves better performance than unsupervised pretraining (\eg SwAV and Mocov2) which are pretrained with 200 epochs. When both pretrained over 300 epochs, SL-MLP shows \textit{better} performance than Byol with \textbf{+1.3 AP}.


\begin{figure*}[t]
    \centering
    \small
    \includegraphics[width=0.93\textwidth]{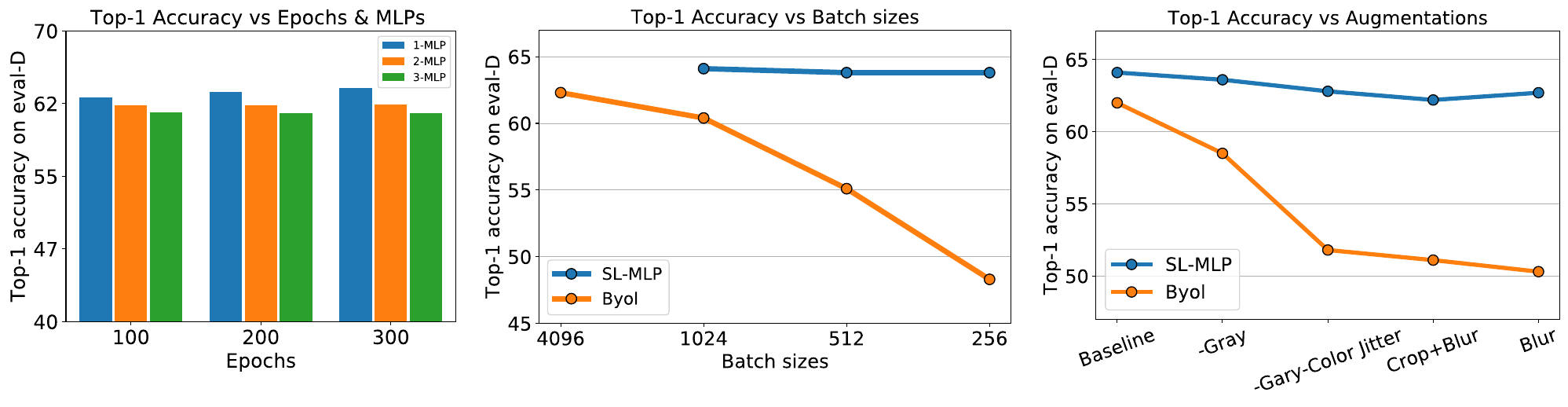}
    \vspace{-.75em}
    \caption{\small  
    (Left to right) \textbf{(a)} Top-1 accuracy with different pretraining epochs and number of MLP projectors. \textbf{(b)} Top-1 accuracy with different batch sizes.
    \textbf{(c)} Top-1 accuracy with different pretraining augmentations.\vspace{-1em}
    }
    \label{fig:epochs&bn&aug}
    \vspace{-.5em}
\end{figure*}

\subsection{Ablation Study}\label{ablation}

\noindent \textbf{Effectiveness of different components in MLP.} In this part, we investigate the influence of different components in the MLP projector. We set the hidden units and output dimension of MLP to be 2048 to retain the dimension of output features the same as SL. Variants are constructed by adding the components incrementally: 
(a) no MLP projector; (b) only Input FC; 
(c) Input FC+BN+output FC; 
(d) Input FC+ReLU+output FC; (e) BN+ReLU. 
All experiments are pretrained on pre-D over 100 epochs. As shown in Tab.~\ref{tab:structural ablation}, SL-MLP achieves the best accuracy among all variants.
We analyze the influence of different components on discriminative ratio $\phi$ on pre-D, Feature Mixtureness $\Pi$, feature redundancy $\mathcal{R}$ qualitatively and quantitatively in Supplementary D.3.
Besides, we also observe an interesting phenomenon on Tab.~\ref{tab:structural ablation}(e) that only inserting a lightweight BN-ReLU also achieves good transfer performance. As this is not our main focus, we will investigate it in future works.

\noindent \textbf{Epochs and layers.}
Fig.~\ref{fig:epochs&bn&aug}(a) shows that adding one MLP projector achieves the optimal transferability.
In addition, larger pretraining epochs benefit the transferability of SL-MLP when one MLP projector is added, but it has little influence when more MLP projectors are used. 

\noindent \textbf{SL-MLP is less sensitive to batch size.}
Most unsupervised methods depend on big mini-batches to train a representation with strong transferability. To investigate the sensitivity of SL-MLP to batch size, we do experiments with batch size from 256 to 4096 for Byol and to 1024 for SL-MLP over 300 epochs. As shown in Fig.~\ref{fig:epochs&bn&aug}(b), the transferability of Byol drops when the batch size decreases. In contrast, the transferability of SL-MLP retains when batch size changes.

\noindent \textbf{SL-MLP is less sensitive to augmentation.}
Unsupervised methods benefit from a broader space of colors and more intensive augmentations during pretraining, which always lead to undesirable degradation when some augmentations are missing. Supervised models trained only with horizontal flipping may perform well~\cite{zhao2021what}. We set Byol's augmentations as a baseline setting for both SL-MLP and Byol. We then compare the robustness on augmentation between SL-MLP and Byol by removing augmentation step by step. Experiments of SL-MLP and Byol are all constructed on their default condition with only augmentations changed. The results are illustrated on Fig.~\ref{fig:epochs&bn&aug}(c). We find that SL-MLP inherits the robustness of SL and shows a little accuracy drop with simple augmentations.



\section{Limitations and Conclusions}
In this paper, we study the transferability gap between supervised and unsupervised pretraining. Based on empirical results, we identify that the MLP projector is a key factor for the good transferability of unsupervised pretraining methods. By adding an MLP projector into supervised pretraining methods, we close the gap between supervised and unsupervised pretraining and even make supervised pretraining better. Our finding is confirmed with extensive experiments on diverse backbone networks and various downstream tasks, including the concept generalization tasks, cross-domain image classifications, and objection detection. While the MLP is a simple design for better transferability, there might exist some straightforward designs on the objective function, which we leave for future work.

\section{Acknowledgment}
This work is supported by the Key R\&D Project of Zhejiang Province (No.2022C01056), Natural Science Foundation of Zhejiang Province (No.LQ21F030017), the National Natural Science Foundation of China (No.62127803) and Hetao Shenzhen-Hong Kong Science and Technology Innovation Cooperation Zone (HZQB-KCZYZ-2021045).

{\small
\bibliographystyle{ieee_fullname}
\bibliography{egbib}
}

\appendix

\begin{onecolumn}
\section{Visualization of Feature Mixtureness}
\label{appendix-visualization-feature-mixtureness}
\begin{figure}[t]
    \centering
    \includegraphics[width=0.9\linewidth]{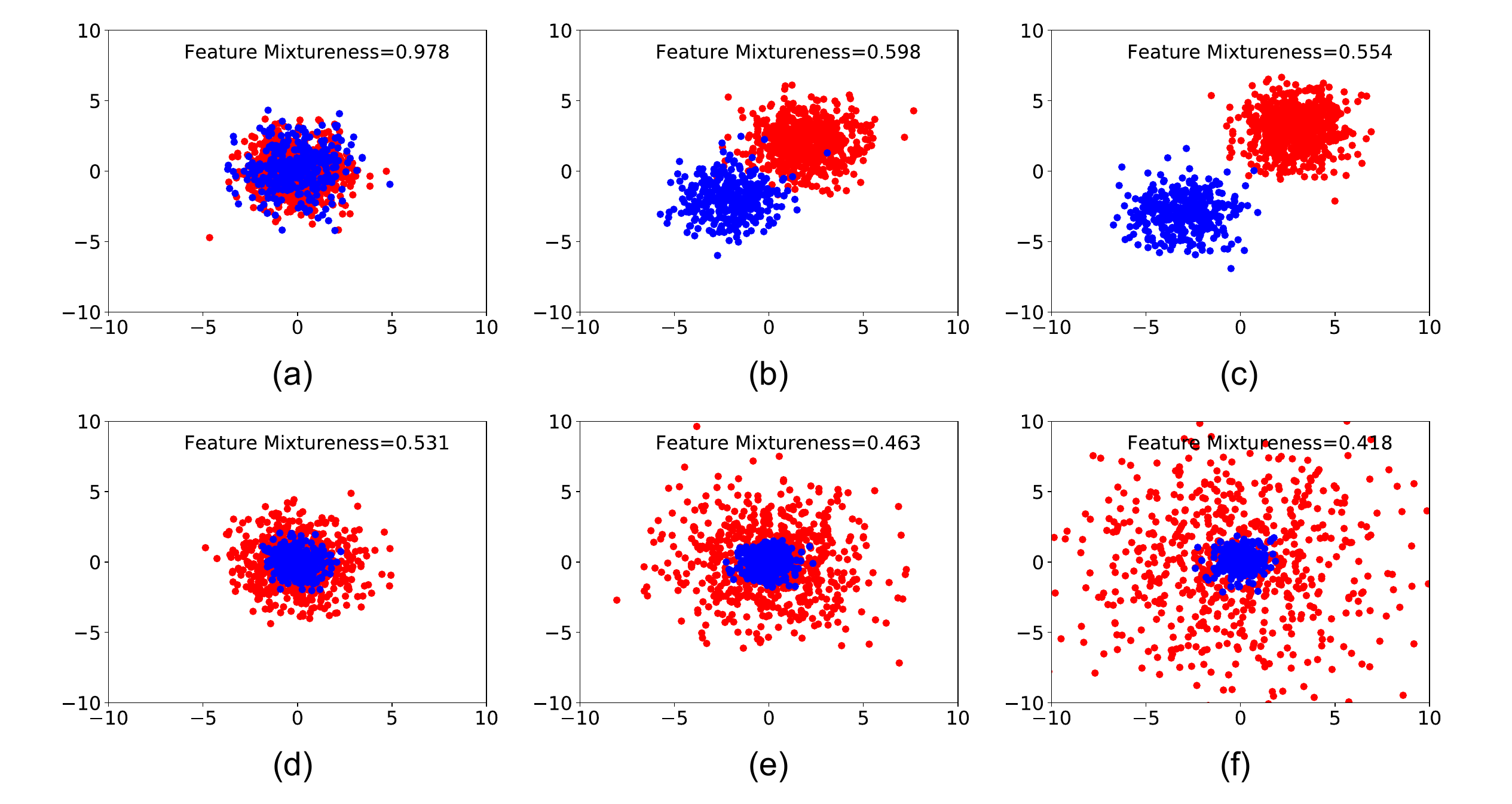}
    \caption{\small{Visualization of Feature Mixtureness with different manually generated feature distribution. Red and blue represent pre-D and eval-D class centers, respectively.  }}
    \label{fig:appendix-visualization-feature-mixtureness}
\end{figure}

We provide an intuitive understanding of the relation between Feature Mixtureness and the feature distribution distance by manually generating two sets of features with different distribution distance. We use red and blue to represent class centers from pre-D and eval-D, respectively. The visualization results are illustrated in Fig.~\ref{fig:appendix-visualization-feature-mixtureness}. From (a) to (c), when the distribution distance between pre-D and eval-D increases, Feature Mixtureness decreases accordingly. When we fix the variance of features in pre-D and gradually enlarge the variance of features in eval-D (from (d) to (f)), Feature Mixtureness will decrease as well. Based on the observations above, we conclude that our Feature Mixtureness can empirically measure the feature distribution distance between pre-D and eval-D.

\section{Visualization of Feature Distribution during Pretraining}
In this section, we provide an illustration to establish an intuition about how intra-class variation and Feature Mixtureness evolve during different pretraining epochs. 

\begin{figure}
    \centering
    \includegraphics[width=.9\linewidth]{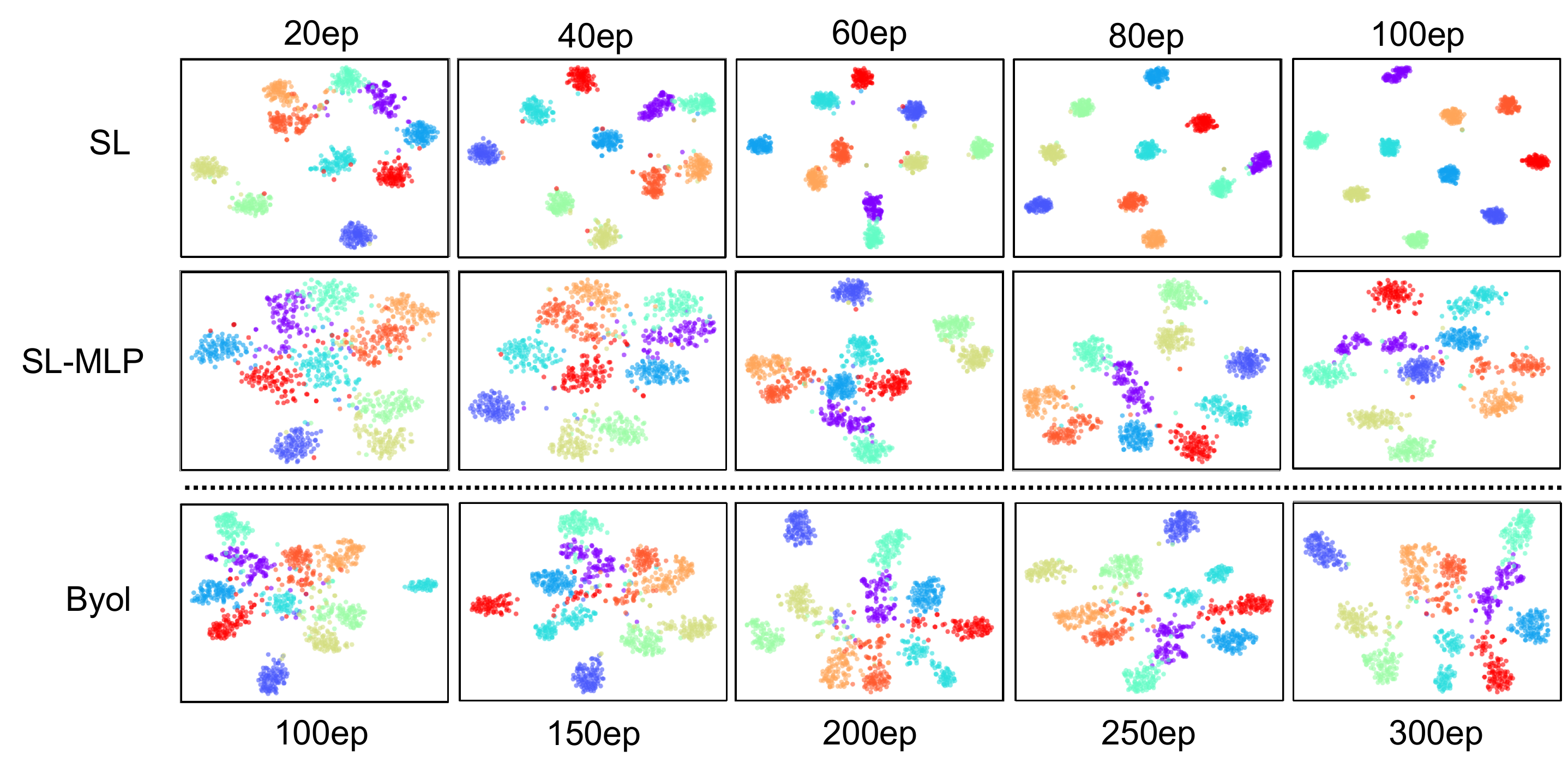}
    \caption{\small{Evolution of intra-class variation of features in pre-D with different epochs. Different colors denote different classes. The intra-class variation of SL will be very small when the pretraining epoch is large enough. Instead, the intra-class variation of SL-MLP and Byol still retains even though the model is pretrained by large epochs.}}
    \label{fig:appe_vis_variation_ep}
\end{figure}

\begin{figure}
    \centering
    \includegraphics[width=.9\linewidth]{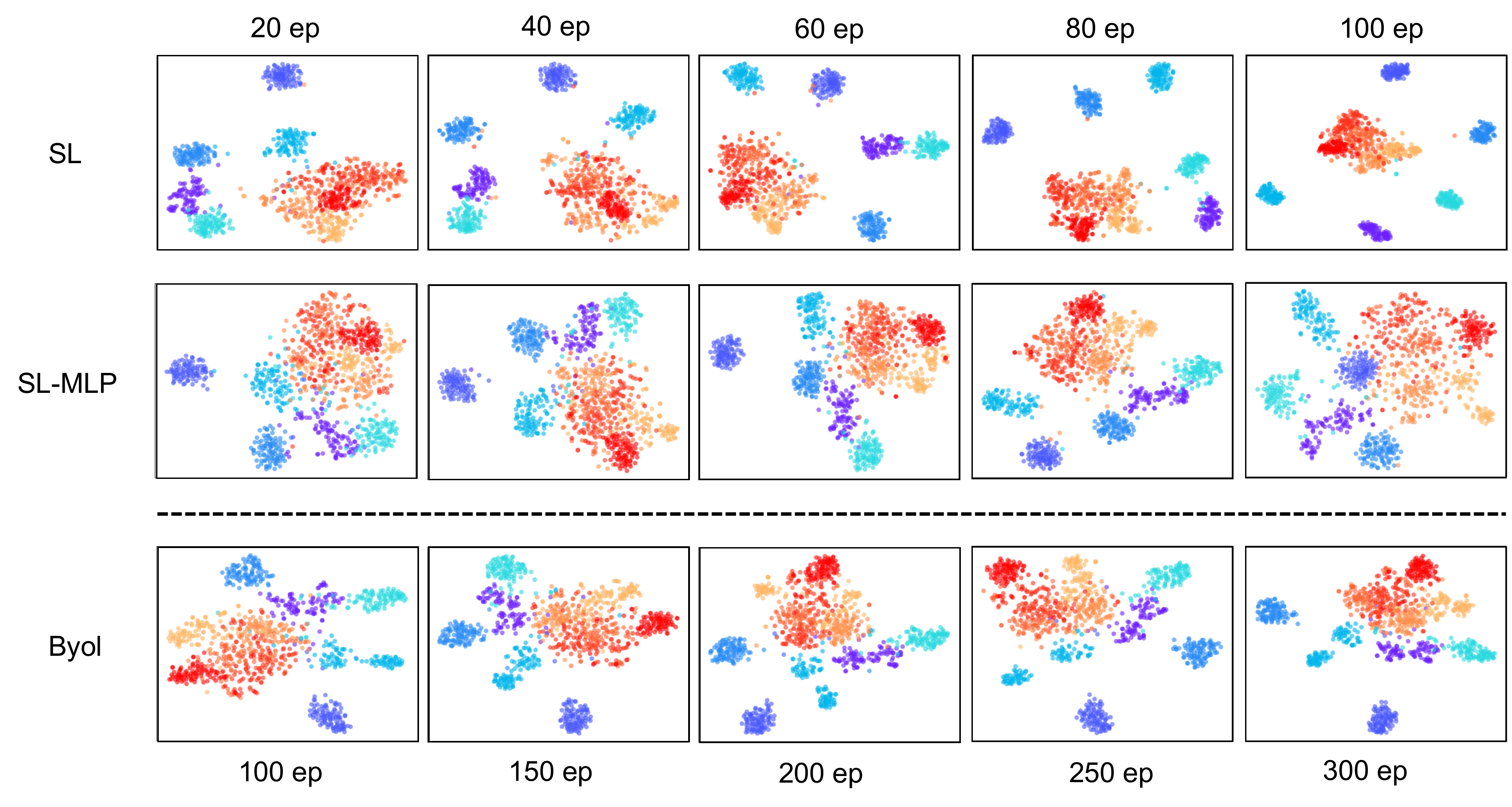}
    \caption{Evolution of Feature Mixtureness between features from pre-D and from eval-D. Cold colors denote features from 5 classes that are randomly selected from pre-D, and warm colors denote features from 5 classes that are randomly selected from eval-D. Feature Mixtureness of SL continuously decrease during pretraining. Alternatively, SL-MLP and Byol keeps a relatively high Feature Mixtureness at large pretraining epochs. }
    \label{fig:appe-vis-connectivity-ep}
\end{figure}

\subsection{Intra-class Variation on pre-D}
\label{appendix-vis-variation-epochs}
We visualize the feature distribution using samples from 10 randomly selected classes in pre-D in Fig.~\ref{fig:appe_vis_variation_ep} to illustrate the evaluation results of the intra-class variation on pre-D. Different colors represent different classes. In SL, the intra-class variation will continuously decrease to a small value with more training epochs. In contrast, the intra-class variance of SL-MLP and Byol retains even though we pretrain the networks at large pretraining epochs. This visualization graphically validates that the MLP projector can enlarge the intra-class variation of features in pre-D.

\subsection{Feature Mixtureness between pre-D and eval-D} 
\label{appendix-vis-connectivity-epochs}
We randomly select features from 5 classes in pre-D and 5 classes in eval-D, and then visualize them by t-SNE in Fig.~\ref{fig:appe-vis-connectivity-ep}. Cold colors represent features from pre-D and warm colors represent features from eval-D. At the early pretraining stage, all methods show high Feature Mixtureness as they cannot well classify images in pre-D. When the training epoch is becoming larger, SL shows lower Feature Mixtureness, which indicates a larger feature distribution distance between pre-D and eval-D. Instead, SL-MLP and Byol remain large Feature Mixtureness when the training epoch is becoming larger, which shows that the feature distribution distance between pre-D and eval-D is not enlarged by Byol and SL-MLP.

\section{Theoretical Analysis of Theory 1}
\label{sec:theoretical_analysis}

\subsection{Proof of Theory 1} 
\begin{proof}
Denote the pretrained feature extractor with the parameters $\mathbf{\theta} $ as $f(\cdot;\mathbf{\theta})$. The softmax function is built upon the feature representation of the backbone $\mathbf{f}_i=f(\mathbf{x}_i;\theta)\in\mathbb{R}^D$, where $\mathbf{x}_i$ is an image, and $D$ is the dimension of features. We compute the class center $\mu(I_j)$ for class $j$ as the mean of the feature embeddings as 
\begin{equation} \label{eq:feature_center}
    \mu(I_j)=\frac{1}{I_j}\displaystyle\sum_{(\mathbf{x}_i,y_i)\in I_j} \displaystyle \mathbf{f}_i,
\end{equation}
where $I_j$ denotes the images in the $j$-th class. Then we define the inter-class distance $D_{inter}(I)$, and intra-class distance $D_{intra}(I)$ on a datatset with $C$ classes as 
 \begin{equation} \label{eq:intra_class}
    D_{inter}(I)=\frac{1}{C(C-1)}\displaystyle\sum_{j=1}^{C}\displaystyle\sum_{k=1,k\neq j}^C \displaystyle || \mu(I_j)-\mu(I_k)||^2,
 \end{equation}
 \begin{equation} \label{eq:inter_class}
    D_{intra}(I)=\frac{1}{C}\displaystyle\sum_{j=1}^{C}(\frac{1}{|I_j|}
     \displaystyle\sum_{(\mathbf{x}_i,y_i)\in I_j} \displaystyle ||\displaystyle \mathbf{f}_i-\mu(I_j)||^2).
 \end{equation}
Substituting Eq.~\ref{eq:feature_center} into Eq.~\ref{eq:intra_class} and Eq.~\ref{eq:inter_class}, we have
\begin{equation} 
\label{eq:pair_inter}
D_{inter}(I)=\frac{1}{C(C-1)}\displaystyle\sum_{j=1}^{C}\displaystyle\sum_{k=1,k\neq j}^C \left(\frac{1}{2|I_j||I_k|}\displaystyle\sum_{(\mathbf{x}_i,y_i)\in I_j}\displaystyle\sum_{(\mathbf{x}_l,y_l) \in I_k} \displaystyle ||\displaystyle \mathbf{f}_i-\displaystyle \mathbf{f}_l||^2\right),
\end{equation}
\begin{equation}
\label{eq:pair_intra}
D_{intra}(I)=\frac{1}{C}\displaystyle\sum_{j=1}^{C}\left(\frac{1}{2|I_j|^2}\displaystyle\sum_{(\mathbf{x}_i,y_i)\in I_j} \displaystyle\sum_{(\mathbf{x}_l,y_l)\in I_j}\displaystyle ||\displaystyle \mathbf{f}_i-\displaystyle \mathbf{f}_l||^2\right).
\end{equation}
Taking expectation to Eq.~\ref{eq:pair_inter} and Eq.~\ref{eq:pair_intra}, for any pair of data $(\mathbf{x}_i,y_i),(\mathbf{x}_l,y_l)\in I$, we have
\begin{equation}
    \mathbb{E}(\displaystyle ||\displaystyle \mathbf{f}_i-\mathbf{f}_l||^2)=
    \begin{cases}
    2D_{intra}(I),y_i=y_l \\
    2D_{inter}(I),y_i \neq y_l
    \end{cases}.
\end{equation}
For ease of analysis, we denote $I^{pre}$, $I^{eval}$ as pre-D and eval-D, respectively. For any pair of data $\displaystyle(\mathbf{x}_i' ,y_i'),(\mathbf{x}_l',y_l') \in I^{eval}$ in eval-D in the same class, \emph{i.e.,} $y'_i=y'_l$, we have
\begin{equation}
 \label{eq:intra_eval_inter_relation}
 \begin{split}
     D_{intra}(I^{eval})&=\frac{1}{2}\mathbb{E}\left(\displaystyle || \mathbf{f}_i'-\mathbf{f}_l'||^2\right)\\
     &=\frac{1}{2}\mathbb{E}\left[
     \displaystyle P(y_i=y_l)2D_{intra}(I^{pre})+
     \displaystyle P(y_i \neq y_l)2D_{inter}(I^{pre}) \right] \\
     &=PD_{intra}(I^{pre})+(1-P)D_{inter}(I^{pre}),
 \end{split}
\end{equation}
where $y_i$ is the label of an image $\mathbf{x}_i$ assigned by the classifier trained on pre-D, and  $\mathbf{f}'=f(\mathbf{x}', \mathbf{\theta})$. Here, $P$ represents the possibility that a pair of images in eval-D that belong to the same class is classified into the same classes in pre-D. 

We denote $\psi(\phi^{-1}(I^{pre}))=D_{inter}(I^{eval})/D_{inter}(I^{pre})$ as the ratio of the model's inter-class distance on eval-D and the model's inter-class distance on pre-D. When the model is optimized on pre-D, its discriminative ratio on pre-D $\phi(I^{pre})$ becomes larger with the increase of $D_{inter}(I^{pre})$ and the decease of $D_{intra}(I^{pre})$. In most cases, $D_{inter}(I^{eval})/D_{inter}(I^{pre})$ is a monotonic decreasing function of $\phi(I^{pre})$, and is a monotonic increasing function of $\phi^{-1}(I^{pre})$, which has been empirically proven by~\cite{liu2020negative}. Mathematically, it can be formulated as 
\begin{equation}
    \label{eq:inscreasing psi}
    \psi(\phi_2^{-1}(I^{pre})) > \psi(\phi_1^{-1}(I^{pre})), \text{ if }  \phi_2^{-1}(I^{pre})>\phi_1^{-1}(I^{pre}).
\end{equation}

By substituting $D_{intra}(I^{eval})=PD_{intra}(I^{pre})+(1-P)D_{inter}(I^{pre})$ (Eq.~\ref{eq:intra_eval_inter_relation}) into the discriminative ratio inequality $\phi_2(I^{eval})<\phi_1(I^{eval})$ given $\phi_2(I^{pre})>\phi_1(I^{pre})$, we have
\begin{alignat}{2}
 \small
     &\phi_2(I^{eval})<\phi_1(I^{eval})\\
     &\iff\frac{D_{inter}^2(I^{eval})}{D_{intra}^2(I^{eval})}<\frac{D_{inter}^1(I^{eval})}{D_{intra}^1(I^{eval})}  \\
     &\iff \frac{D_{inter}^2(I^{eval})}{PD_{intra}^2(I^{pre})+(1-P)D_{inter}^2(I^{pre})}<\frac{D_{inter}^1(I^{eval})}{PD_{intra}^1(I^{pre})+(1-P)D_{inter}^1(I^{pre})}, \\
     &\iff P < \frac{\frac{D_{inter}^1(I^{eval})}{D_{inter}^1(I^{pre})}-\frac{D_{inter}^2(I^{eval})}{D_{inter}^2(I^{pre})}}{\frac{D_{inter}^1(I^{eval})}{D_{inter}^1(I^{pre})}\cdot\left(1-\frac{D_{intra}^2(I^{pre})}{D_{inter}^2(I^{pre})}\right)-\frac{D_{inter}^2(I^{eval})}{D_{inter}^2(I^{pre})}\cdot\left(1-\frac{D_{intra}^1(I^{pre})}{D_{inter}^1(I^{pre})}\right)}, \\
     &\iff P < \frac{\psi(\phi^{-1}_1(I^{pre}))-\psi(\phi^{-1}_2(I^{pre}))}{\psi(\phi^{-1}_1(I^{pre}))\left(1-\phi_2^{-1}(I^{pre})\right)-\psi(\phi^{-1}_2(I^{pre}))\left(1-\phi_1^{-1}(I^{pre})\right)}, \\
     &\iff P<\frac{1}{1-\phi^{-1}_1(I^{pre})+\frac{\phi_2^{-1}(I^{pre})-\phi_1^{-1}(I^{pre})}{\psi(\phi_2^{-1}(I^{pre}))-\psi(\phi_1^{-1}(I^{pre}))}\psi(\phi^{-1}_1(I^{pre}))}, \\
     &\iff P<\frac{1}{1-\phi^{-1}_1(I^{pre})+r\psi(\phi^{-1}_1(I^{pre}))}, \\
     &\iff r\psi(\phi^{-1}_1(I^{pre})) - \phi^{-1}_1(I^{pre}) < P^{-1}-1, \\
     &\iff \frac{d\phi^{-1}_1(I^{pre})}{d\psi(\phi^{-1}_1(I^{pre}))}\psi(\phi^{-1}_1(I^{pre}))-\phi^{-1}_1(I^{pre}) < P^{-1}\!-\!1, \\
      \label{eq:derive_relation}
     &\iff \frac{d\phi^{-1}(I^{pre})}{P^{-1}-1+\phi^{-1}(I^{pre})} < \frac{1}{\psi(\phi^{-1}(I^{pre}))}d\psi(\phi^{-1}(I^{pre})),
\end{alignat}
where 
\begin{alignat}{2}
        \small
    r&= \frac{\phi_2^{-1}(I^{pre})-\phi_1^{-1}(I^{pre})}{\psi(\phi_2^{-1}(I^{pre}))-\psi(\phi_1^{-1}(I^{pre}))}\\
&\approx \frac{d \phi^{-1}(I^{pre}) }{d \psi(\phi^{-1}(I^{pre}))}, \text{ when } \phi_2^{-1}(I^{pre})-\phi_1^{-1}(I^{pre})\rightarrow 0.
\end{alignat}


We take integration of Eq.~\ref{eq:derive_relation} as
\begin{alignat}{2}
\small
 &\iff \int_0^{\phi^{-1}(I^{pre})} \frac{d\phi^{-1}(I^{pre})}{P^{-1}-1+\phi^{-1}(I^{pre})} < \int_{\psi(0)}^{\psi(\phi^{-1}(I^{pre}))}\frac{1}{\psi(\phi^{-1}(I^{pre}))}d\psi(\phi^{-1}(I^{pre})), \\  
 &\iff \ln\left[\phi^{-1}(I^{pre})+P^{-1}-1\right] < \ln\left[\psi(\phi^{-1}(I^{pre})))\right] + \ln \left(\frac{P^{-1}-1}{\psi(0)}\right), \\
 &\iff \phi^{-1}(I^{pre})+P^{-1}-1<\psi(\phi^{-1}(I^{pre}))\frac{P^{-1}-1}{\psi(0)}, \\
 &\iff \phi^{-1}(I^{pre}) < 1-P^{-1}+\psi(\phi^{-1}(I^{pre}))\frac{P^{-1}-1}{\psi(0)}, \\
 &\iff \phi^{-1}(I^{pre}) < (\frac{\psi(\phi^{-1}(I^{pre}))}{\psi(0)}-1)(P^{-1}-1)\\
 &\iff \phi(I^{pre}) > t
\end{alignat}
where the threshold $t$ is defined as
\begin{equation}
    \label{eq:defination_t}
    t=\left[(\frac{\psi(\phi^{-1}(I^{pre}))}{\psi(0)}-1)(P^{-1}-1)\right]^{-1}. 
\end{equation}

According to Formulation~\ref{eq:inscreasing psi},  $\psi(\phi^{-1}(I^{pre}))>{\psi(0)}$ because $\phi^{-1}(I^{pre})>0$.
Therefore, $\frac{\psi(\phi^{-1}(I^{pre}))}{\psi(0)}-1>0$, which means that increasing $P$ will lead to increasing the threshold $t$. 
\end{proof}

\subsection{Analysis of P}
In the following, we explain how $P$ in Equation~\ref{eq:intra_eval_inter_relation} can be theoretically computed, and how $P$ negatively relates to the feature distribution distance briefly.

\subsubsection{Computational Method of P} Given a fixed backbone pretrained $f(\cdot; \mathbf{\theta})$ on pre-D, we denote the classifier trained by pre-D as $\mathbf{W}=(\mathbf{w}_1, \mathbf{w}_2, ..., \mathbf{w}_{C^{pre}})$. The possibility of an image $\mathbf{x}$ of the class $j$ in eval-D classified by the classifier $\mathbf{W}$ into the class $k$ in pre-D can be defined as
\begin{equation}
    P_{jk}=\frac{1}{|I_j^{eval}|}\sum_{(\mathbf{x}_i,y_i) \in I_j^{eval}} \frac{\exp(\mathbf{w}_k \cdot f(\mathbf{x};\mathbf{\theta}))}{\sum_{k'=1}^{C^{pre}}\exp(\mathbf{w}_{k'} \cdot f(\mathbf{x};\mathbf{\theta}))},
\end{equation}
where $|I_j^{eval}|$ denotes the number of images in the $j$-th class in eval-D. Then the probability of a pair of samples in the same class $j$ in eval-D classified into the same class in eval-D is 
\begin{equation}
    P_j = \sum_{k=1}^{C^{pre}}P^2_{jk}.
\end{equation}
The average probability of $P_j$ is
\begin{equation}
     P = \frac{1}{C^{eval}}\sum_{j=1}^{C^{eval}}P_j.
\end{equation}

\subsubsection{P is Negatively Related to the Feature Distribution Distance} 
In this part, we only use two extreme cases to briefly analyze the relation between $P$ and the feature distribution distance.

Specifically, we first deduce the upper bound and the lower bound of $P$. We find that the upper bound is reached when the feature distribution distance between pre-D and eval-D is extremely small, and the lower bound is reached when the feature distribution distance between pre-D and eval-D is extremely large, which indicates $P$ is negatively related to the feature distribution distance.

For the upper bound of $P$,
\begin{alignat}{2}
    P &= \frac{1}{C^{eval}}\sum_{j=1}^{C^{eval}}P_j \\ &=\frac{1}{C^{eval}}\sum_{j=1}^{C^{eval}}\sum_{k=1}^{C^{pre}}P^2_{jk} \\
    \label{eq:cauchy_inequality_P}
    &\leq \frac{1}{C^{eval}} \sum_{j=1}^{C^{eval}}\left(\sum_{k=1}^{C^{pre}}P_{jk}\right)^2 \\
    &=\frac{1}{C^{eval}} \sum_{j=1}^{C^{eval}} 1 \\
    &=1, 
\end{alignat}
where Inequality~\ref{eq:cauchy_inequality_P} is derived by Cauchy Schwarz Inequality~\cite{wu2009various}, and if and only if $P_{jk}=1$ and $P_{jk'}=0$ for $\forall k' \neq k$, $P$ reaches its upper bound 1.

For the lower bound of $P$,
\begin{alignat}{2}
    P &= \frac{1}{C^{eval}}\sum_{j=1}^{C^{eval}}P_j \\ &=\frac{1}{C^{eval}}\sum_{j=1}^{C^{eval}}\sum_{k=1}^{C^{pre}}P^2_{jk} \\
    \label{eq:fundamental_inequality_P}
    &\geq \frac{1}{C^{eval}}\sum_{j=1}^{C^{eval}}\frac{1}{C^{pre}} \left(\sum_{k=1}^{C^{pre}}P_{jk}\right)^2 \\
    &=\frac{1}{C^{eval}} \sum_{j=1}^{C^{eval}} \frac{1}{C^{pre}} \\
    &=\frac{1}{C^{pre}}, 
\end{alignat}
where Inequality~\ref{eq:fundamental_inequality_P} is derived by Fundamental Inequality~\cite{beckenbach1961introduction}, and if and only if $P_{jk}=\frac{1}{C^{pre}}$ for $\forall k \in [1, C^{pre}]$, $P$ reaches its lower bound $\frac{1}{C^{pre}}$.

\underline{\emph{Analysis on Small Feature Distribution Distance between pre-D and eval-D.}} 
When pre-D and eval-D have small feature distribution distance, a pair of two images $(\mathbf{x}_m,y'_m)$ and $(\mathbf{x}_n, y'_n)$ belong to the same class $j$  in eval-D, \emph{i.e.,} $y'_m=y'_n$ will be classified to the same class $k$ in pre-D when classified by $\mathbf{W}$ with high confidence. That is, only $P_{jk}$ will have high confidence close to 1 and $P_{jk^{'}},$ $\forall k' \neq k$ will be close to 0, which is similar to the condition when $P$ reaches its upper bound.

\underline{\emph{Analysis on Large Feature Distribution Distance between pre-D and eval-D.}} When pre-D and eval-D have large feature distribution distance, a pair of two images $(\mathbf{x}_m,y'_m)$ and $(\mathbf{x}_n, y'_n)$ belong to the same class in eval-D, \emph{i.e.,} $y'_m=y'_n$ will be randomly classified to the classes in pre-D using $\mathbf{W}$. Mathematically, $P_{jk} \approx \frac{1}{C^{pre}}$, which is similar to the condition when $P$ reaches its lower bound.

Based on the analysis above, we can conclude that $P$ is negatively related to feature distribution distance, and larger $P$ often means less feature distribution distance.

\section{MLP components}
In this section, we provide the detailed analysis about how each component of the MLP projector influences the intra-class variation (represented by discriminative ratio $\phi^{pre}$) on pre-D, Feature Mixtureness $\Pi$ between pre-D and eval-D, and feature redundancy $\mathcal{R}$. Based on SL which does not include MLP, we ablate the structure of the MLP projector by adding the input fully connected layer, the output fully connected layer, the batch normalization layer and the ReLU layer incrementally. The input fully connected layer and the output fully connected layer are both set to have hidden units of 2048 and output dimensions of 2048 to keep same output feature dimensions as SL. All experiments are pretrained over 100 epochs. Testing results of the discriminative ratio on pre-D, Feature Mixtureness $\Pi$ and feature redundancy $\mathcal{R}$ are illustrated in Tab.~\ref{tab:quan-mlp-design}.

\subsection{Visualization of intra-class variation}
We randomly select features from 10 classes in pre-D and visualize their intra-class variation in Fig.~\ref{fig:appe_vis_MLP_component_variation}. Different colors denote features from different classes. We specify the components in the MLP projector below each visualization image. Comparing (a) with (b), we can see that adding a fully connected layer can slightly enlarge intra-class variation, which indicates that linear transformation helps transferability marginally. Instead, comparing (a-b) with (c-e), we can observe that the batch normalization layer and the ReLU layer are important components in the MLP projector, which can significantly enlarge the intra-class variation in the feature space of pre-D. In general, comparing SL-MLP with (a-e), we can conclude that all components in MLP projector help enlarge the intra-class variation of features in pre-D while the batch normalization layer and the ReLU layer play the most important roles.

\begin{figure}[t]
\begin{floatrow}
\floatbox{figure}[0.47\textwidth]
{\caption{\small{Visualization of intra-class variation by different components. We randomly select 10 classes in pre-D. Different colors denote different classes. Comparing (a) wth (b), we can see the fully-connected layer can slightly help enlarge the intra-class variation. Comparing (a-b) and (d-e), we can observe the batch normalization layer and the ReLU layer can significantly enlarge the intra-class variation in the feature space. In general, all components in the MLP layer is beneficial to enlarge intra-class variation, which proves their effectiveness in enhancing transferaiblity of pretraining models.}}
 \label{fig:appe_vis_MLP_component_variation}}
{\includegraphics[width=\linewidth]{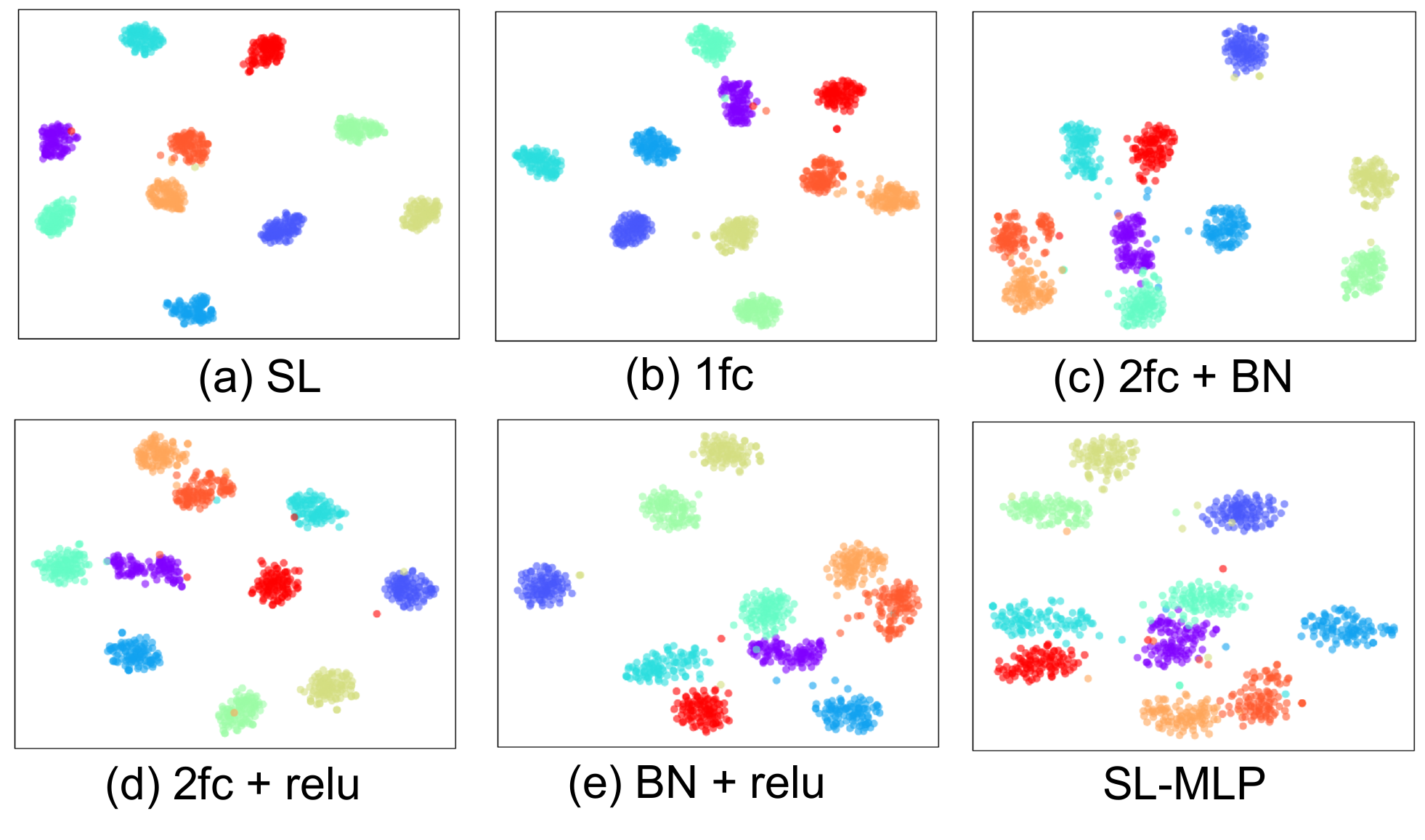}
\centering
}
\floatbox{figure}[.47\textwidth]
{\caption{\small{Visualization of Feature Mixtureness of features pretrained by different MLP components. Different colors denote different classes. Points with cold colors denote the features from pre-D, and points with warm colors denote the features from eval-D. Comparing (c-d) with (a-b), we can see that adding BN and ReLU can increase Feature Mixtureness between pre-D and eval-D. Comparing (e) with (a-d), we can conclude that BN and ReLU play the main roles in the MLP projector as (e) shows larger Feature Mixtureness. An MLP projector with all components achieves the largest Feature Mixtureness.}}
 \label{fig:appendix-vis-mlp-component-connectivity}}
{\centering
    \includegraphics[width=\linewidth]{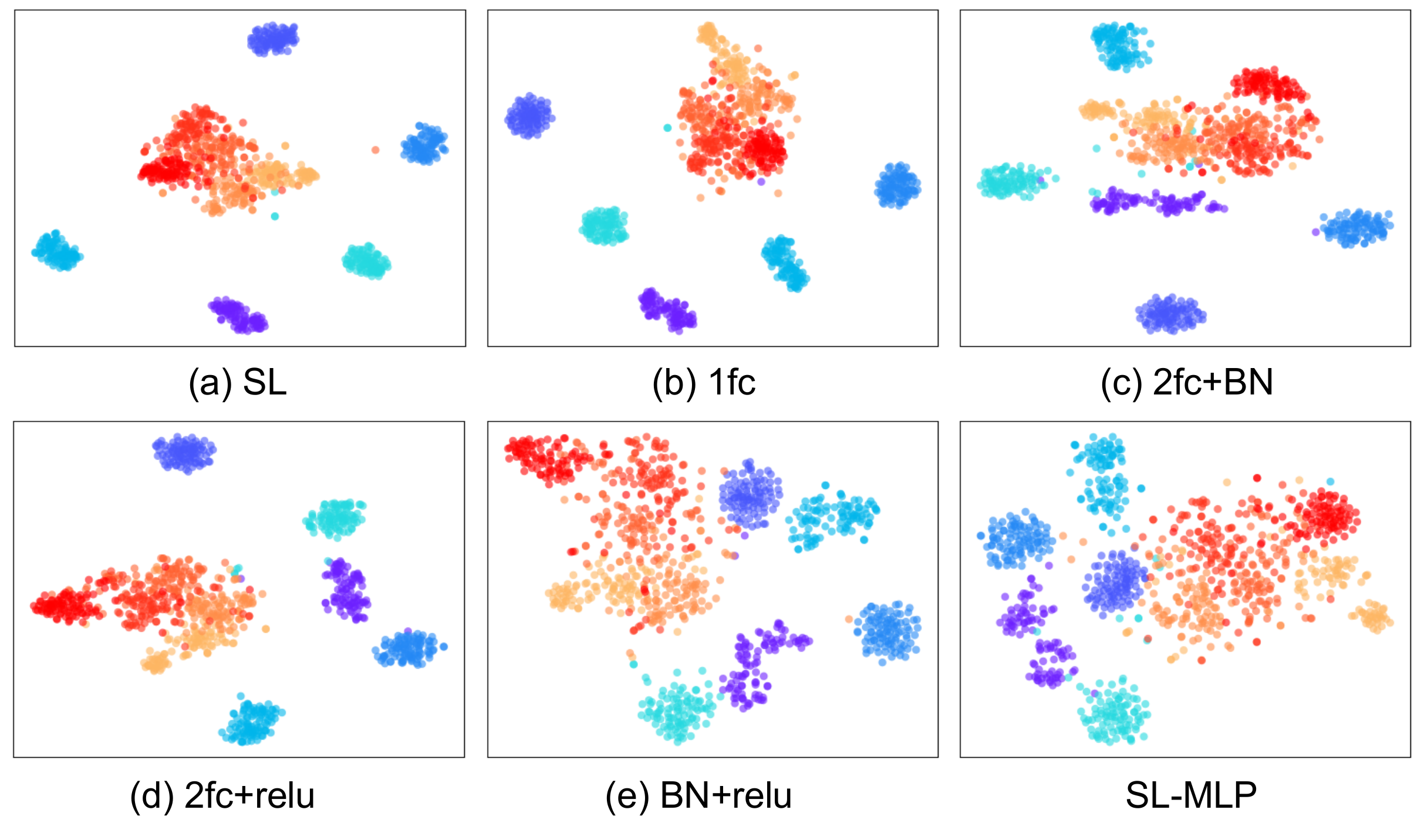}}
\end{floatrow}
\end{figure}

\subsection{Visualization of Feature Mixtureness}

We randomly select features from 5 classes in pre-D and 5 classes in eval-D to visualize Feature Mixtureness with different MLP components. The results are summarized in Fig.~\ref{fig:appendix-vis-mlp-component-connectivity}. The features with cold colors come from pre-D, the features with warm colors come from eval-D. Comparing (a) and (b), we can see adding a fully connected layer can hardly increase Feature Mixtureness between pre-D and eval-D. Comparing (c-d) with (b), we can conclude that the batch normalization layer and the ReLU layer can increase Feature Mixtureness between pre-D and eval-D. Comparing (b-d) with (e), we can summarize that the batch normalization and the ReLU layer are the most important components. A batch normalization layer with a ReLU layer can significantly increase Feature Mixtureness between pre-D and eval-D, which has already been similar to Feature Mixtureness when the MLP projector has the complete architectural.


\subsection{Quantitative Analyse of MLP components}
With the discriminative ratio $\phi^{pre}$, Feature Mixtureness $\Pi$ and feature redundancy $\mathcal{R}$ defined in Sec.~\ref{sec-finding-SLMLP}, we quantitatively examine the effect of different components in the MLP projector. The results are presented in Tab.~\ref{tab:quan-mlp-design}. Firstly, the fully connected layer has little influence on three metrics. Comparing (a) and (b), when adding a fully connected layer, the model shows slight improvement on Feature Mixtureness and feature redundancy, and slight decrease of the discriminative ratio on pre-D. Second, non-linear layer brings considerable improvements. Comparing (b) to (d), we can summarize that incrementally adding a ReLU, a batch normalization layer can increase Feature Mixtureness, reduce discriminative ratio, which could improve transferability of the pretrained model. Specifically, the ReLU layer brings a little improvement on feature redundancy. Comparing (a,b) with (c,e), we can conclude that BN not only reduces the discriminative ratio on pre-D, but also increases Feature Mixtureness. BN has a significant influence on future redundancy, which reduces feature redundancy by 50\% (from 0.0671 to 0.0369). Last but not least, the combination of all components achieves the best transferability with the lowest feature redundancy, the highest Feature Mixtureness and a relatively large intra-class variation.

\label{Appendix: quan-mlp-design}

\begin{table*}[tbp]
  \centering
  \small
  \caption{\small{Quantitative analysis of structural design of inserted MLP, including discriminative ratio on pre-D, Feature Mixtureness $\Pi$ and feature redundancy $\mathcal{R}$. (b-e) denote experiments in which different components are added on the SL baseline (a). When incrementally adding components of the MLP into SL, the distriminative ratio on pre-D and feature redundancy will decrease while the Feature Mixtureness will increase.}  }
    \begin{tabular}{ccccccccc}
    \toprule
    \multicolumn{1}{c}{\multirow{2}[4]{*}{Exp}} & \multicolumn{4}{c}{Components} & \multicolumn{1}{c}{\multirow{2}[4]{*}{Top-1}} & \multicolumn{1}{c}{\multirow{2}[4]{*}{$D^{pre}_{inter}/D^{pre}_{intra}$}} & \multicolumn{1}{c}{\multirow{2}[4]{*}{$\Pi(\uparrow)$}} & \multicolumn{1}{c}{\multirow{2}[4]{*}{$\mathcal{R}(\downarrow)$}} \\
\cmidrule{2-5}          & Input FC & BN    & ReLU  & Output FC &       &       &       &  \\
    \midrule
    (a)     &       &       &       &       & 55.9  & 2.034  & 0.515  & 0.0776  \\
    (b)    & \checkmark     &       &       &       & 56.6  & 1.505  & 0.679  & 0.0671  \\
    (c)    &  \checkmark     &  \checkmark     &       &  \checkmark    & 61.0    & 1.269  & 0.870  & 0.0369  \\
    (d)    &  \checkmark     &       &  \checkmark    &  \checkmark     & 60.1  & 1.362  & 0.804  & 0.0654  \\
    (e)    &       &  \checkmark     &  \checkmark     &       & 60.5  & 1.045  & 0.846  & 0.0369  \\
    \midrule
    SL-MLP &  \checkmark     &  \checkmark     &  \checkmark     &  \checkmark     & 62.5  & 1.124  & 0.871  & 0.0351  \\
    \bottomrule
    \end{tabular}%
  \label{tab:quan-mlp-design}%
\end{table*}%

\section{Concept Generalization Task with Small Semantic Gap} \label{appendix:little_semantic_gap}

Following~\cite{sariyildiz2021concept}, to investigate how semantic difference between pre-D and eval-D can influence the transfer results on \emph{Concept generalization task}, we randomly choose 652 classes as pre-D and 348 classes as eval-D from ImageNet-1K to establish a benchmark where pre-D and eval-D have small semantic gap. We denote the setting where pre-D and eval-D are constructed as Sec.~3.1 as large semantic gap setting (dubbed as \emph{semantic}), and denote the setting where pre-D and eval-D are randomly selected as small semantic gap setting (dubbed as \emph{random}). 



\begin{figure}[t]
\begin{floatrow}
\floatbox{figure}[0.48\textwidth]
{\caption{\small{Visualization of Feature Mixtureness between pretraining dataset (pre-D) and evaluation dataset (eval-D). Different colors denote different classes. Classes in pre-D are denoted by cold colors, and classes in eval-D are denoted by warm colors. Comparing (a,c,e) and (b,d,f), we can conclude that large semantic gap between pre-D and eval-D will lead to small Feature Mixtureness between pre-D and eval-D. Comparing (b) and (d-f), we can observe that the MLP projector can increase Feature Mixtureness between pre-D and eval-D, and can bridge the semantic gap between pre-D and eval-D.}}
 \label{fig:appe_vis_semantic_random_gap}}
{\includegraphics[width=\linewidth]{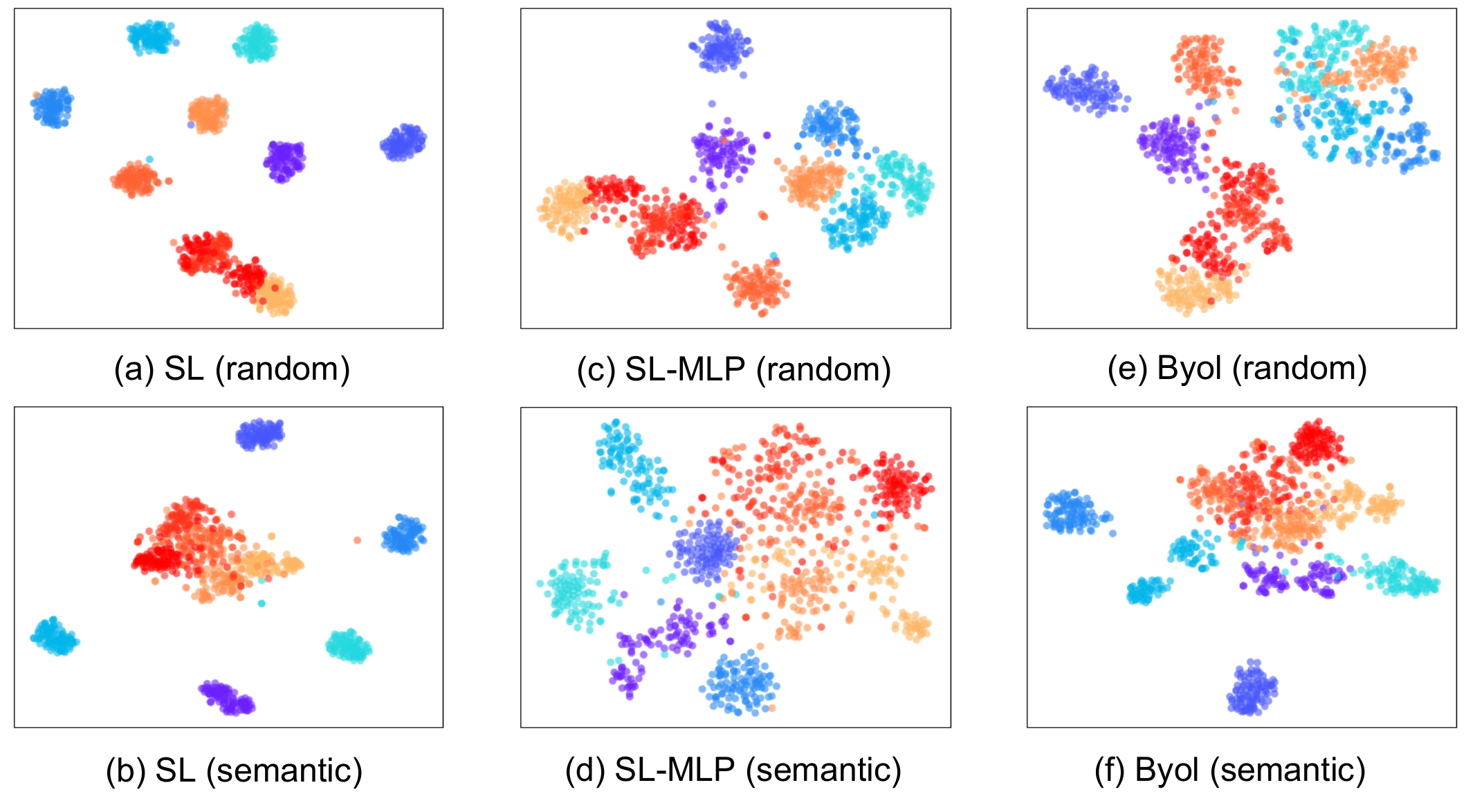}
\centering
}
\floatbox{figure}[.46\textwidth]
{\caption{\small{Linear evaluation accuracy on eval-D with small semantic gap. Following ~\cite{he2019rethinking,grill2020bootstrap}, we pretrain SL, SL-MLP, and Byol on randomly chosen pre-D for 300 epochs. Compare the transfer performance on 300 epochs, SL shows a comparable transferability with Byol and SL-MLP when the semantic gap between pre-D and eval-D is small. In addition, unlike what we observe in Fig.~\ref{fig:Stage-wise-top1-discriminative}(b), no performance drop during the last epochs appear. SL-MLP has a similar performance with SL from 60 to 240 epochs while has a consistently better performance on 300 epochs.}}
 \label{fig:appendix-transfer-random-concept}}
{\centering
    \includegraphics[width=.95\linewidth]{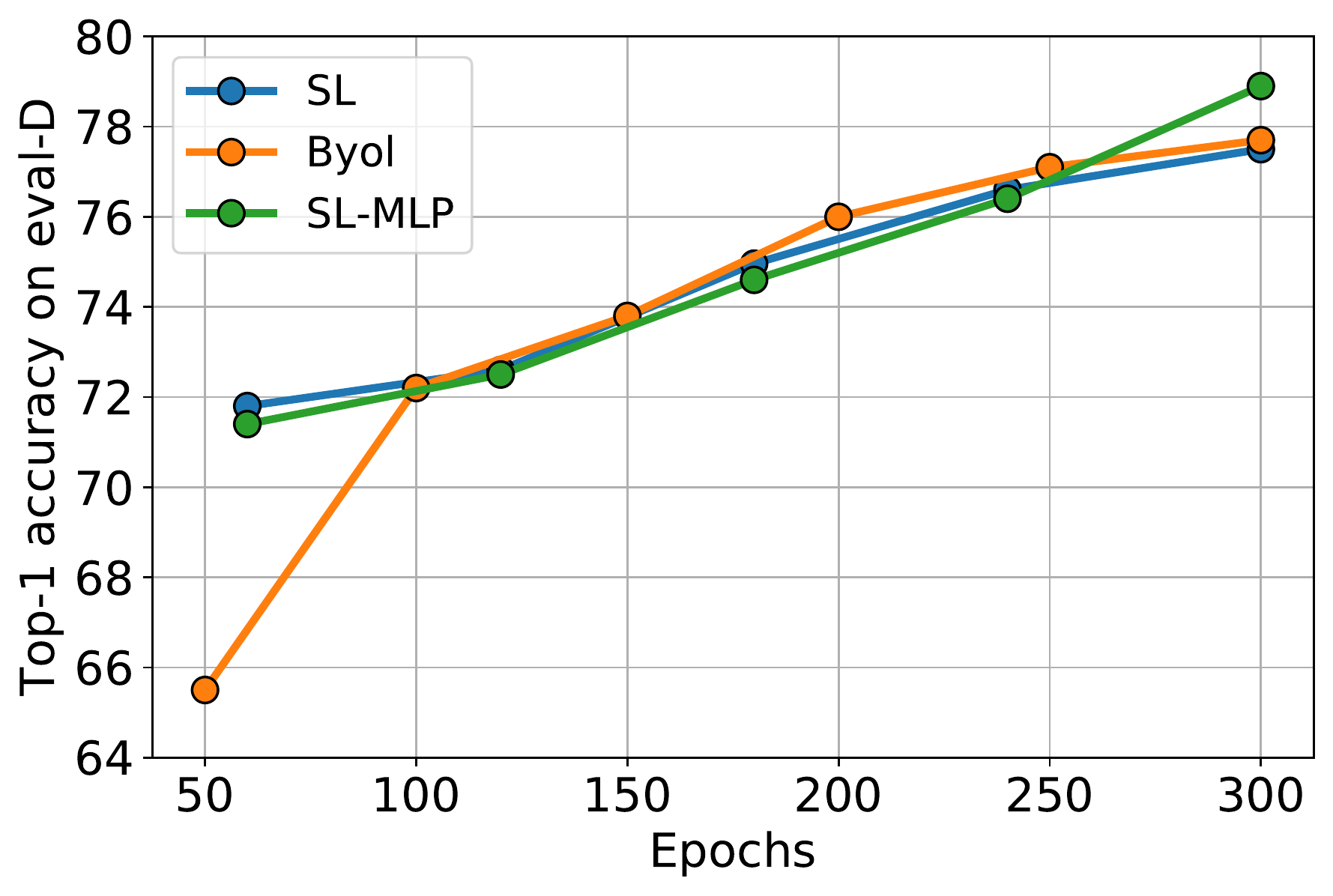}
    \vspace{-0.4em}}
\end{floatrow}
\end{figure}

\subsection{Visualization of Feature Mixtureness}
We visualize features from pre-D and eval-D in small semantic gap setting and large semantic gap setting in Fig.~\ref{fig:appe_vis_semantic_random_gap}.
Specifically, SL (random), SL-MLP (random), and Byol (random) denote feature visualization of SL, SL-MLP, and Byol pretraining on the benchmark where pre-D and eval-D are randomly chosen. SL (semantic), SL-MLP (semantic), and Byol (semantic) denote feature visualization of SL, SL-MLP, and Byol pretraining on the benchmark where pre-D and eval-D are split according to semantic difference in WordNet, which is the same as Sec.~\ref{sec-finding-SLMLP}. Our findings are two-fold. First, comparing with (a), (c), (e), pre-D features in (b), (d), (f) have large Feature Mixtureness, which indicates semantic difference influences the feature distribution distance between pre-D and eval-D in the feature space. Second, comparing (b) with (d), we find that Feature Mixtureness between pre-D and eval-D is enlarged by adding an MLP projector, which indicates that the MLP projector can significantly mitigate the feature distribution distance between pre-D and eval-D. 

\subsection{Quantitative Results} 
We first pretrain all the models on pre-D over 300 epochs, then examine linear evaluation results on eval-D. Our findings are three-fold. Firstly, compare the top-1 accuracy on 300 epochs, SL shows a comparable transferability with Byol and SL-MLP when the semantic gap between pre-D and eval-D is small. Second, unlike what we observe in Fig.~\ref{fig:Stage-wise-top1-discriminative}(b), no performance drop during the last epochs appears, which indicates that the intra-class variation of SL is not above the threshold (defined in Sec.~\ref{sec:theoretical_analysis_sl_mlp}) when pre-D and eval-D have a small semantic gap. Third, SL-MLP has a similar performance with SL from 60 to 240 epochs while has a consistently \textit{better} performance on 300 epochs, which verifies the effectiveness of the added MLP projector.

\section{Replacing Softmax with Cosine-Softmax}
In order to prove that our findings can be compatible with different loss functions, we replace the softmax cross-entropy loss with the cosine-softmax cross-entropy loss in the pretraining stage. Specifically, the cosine-softmax cross-entropy loss is defined as
\begin{equation}
    \mathcal{L}_{\text{cos}}(\mathbf{x}_i, y_i) = - \log \frac{\exp(\beta \cdot \cos(\mathbf{w}_{y_i}, f(\mathbf{x}_i)))}{\sum_{i=j}^C \exp(\beta\cdot \text{cos}(\mathbf{w}_j, f(\mathbf{x}_i)))},
\end{equation}
where $\mathbf{w}_i$ is the $i$-th class prototype, $\beta$ is the scale factor. Accordingly, we add an MLP projector before the classifier to construct cosine-softmax-mlp cross-entropy loss, \emph{i.e.,}
\begin{equation}
    \mathcal{L}_{\text{cos-mlp}}(\mathbf{x}_i, y_i) = - \log \frac{\exp(\beta \cdot \cos(\mathbf{w}_{y_i}, g(f(\mathbf{x}_i))))}{\sum_{i=j}^C \exp(\beta\cdot \text{cos}(\mathbf{w}_j, g(f(\mathbf{x}_i))))},    
\end{equation}
where $\mathbf{w}_i$ is the $i$-th class prototype, $\beta=30$ is the scale factor. We train for 100 epochs with a warm-up of 10 epochs and cosine decay learning schedule using the SGD optimizer. The base learning rate is set to 0.4. Weight decay of $10^{-4}$ is applied during pretraining. 
We report the top-1 accuracy on eval-D in Tab.~\ref{tab:cosine_softmax}. The results illustrate that when the model pretrained by cosine-softmax cross-entropy loss, adding an MLP projector can also facilitate transferability of supervised pretraining methods.

\begin{table}[t]
\centering
\caption{\small{Top-1 linear evaluation accuracy on eval-D when pretraining the model on pre-D by cosine-softmax cross-entropy loss.}}
\begin{tabular}{ccc}
\toprule
epoch & cos  & cos-mlp  \\ 
\midrule
20    & \textbf{47.1} & 45.0       \\
40    & 47.8 & \textbf{49.6}     \\
60    & 50.9 & \textbf{52.6}     \\
80    & 53.5 & \textbf{56.5}     \\
100   & 53.7 & \textbf{59.0}    \\ \bottomrule
\end{tabular}
\label{tab:cosine_softmax}
\end{table}

\section{Visualize Convolution Channels by Optimization}
\label{appendix:visualize layers}
According to~\cite{zhao2021what} and~\cite{asano2019critical}, transfer performance is largely unaffected by the high-level semantic content of the pretraining data. To investigate that whether adding an MLP projector can influence what the convolution channels can learn. By using the method proposed in~\cite{olah2017feature}, we visualize the maximum response of convolution channels in layer 4 of ResNet50 (seen in Fig.~\ref{fig:main_stagewise_eval_illustrate}) pretrained with methods without-MLP, \emph{i.e.} SL, Mocov1, and Byol w/o MLP, and methods with-MLP, \emph{i.e.} SL-MLP, Mocov1 w/ MLP, and Byol. Specifically, given a backbone with fixed parameters $\mathbf{\theta}$ as $f(\cdot;\mathbf{\theta})$, we denote the parameters before the convolution channel $j$ as $f(\cdot;\mathbf{\theta}_{j})$, we optimize the most representative sample $\mathbf{x}_i$ of the convolution channel $j$ by maximizing the output logits $f(\mathbf{x};\mathbf{\theta}_{j})$, \emph{i.e.,} $\mathbf{x}_i = argmax_{\mathbf{x}}(f(\mathbf{x};\mathbf{\theta}_{j}))$, where $\mathbf{x}$ is optimized from a random initialized image $\mathbf{x}_0$.

As shown in Fig.~\ref{fig:appe_layer_vis}, methods without-MLP (Mocov1, Byol w/o MLP, SL) learn more knowledge about animals from pre-D, highlighted by red rectangles. This is due to that we select classes of organisms to construct pre-D. 
Instead, we find that methods with-MLP (Mocov1 w/ MLP, Byol, SL-MLP) learn more texture information. According to~\cite{zhao2021what}, high-level semantic information is less critical to transfer learning, which explains effectiveness of the MLP.

\begin{figure*}
    \centering
    \includegraphics[width=.9\linewidth]{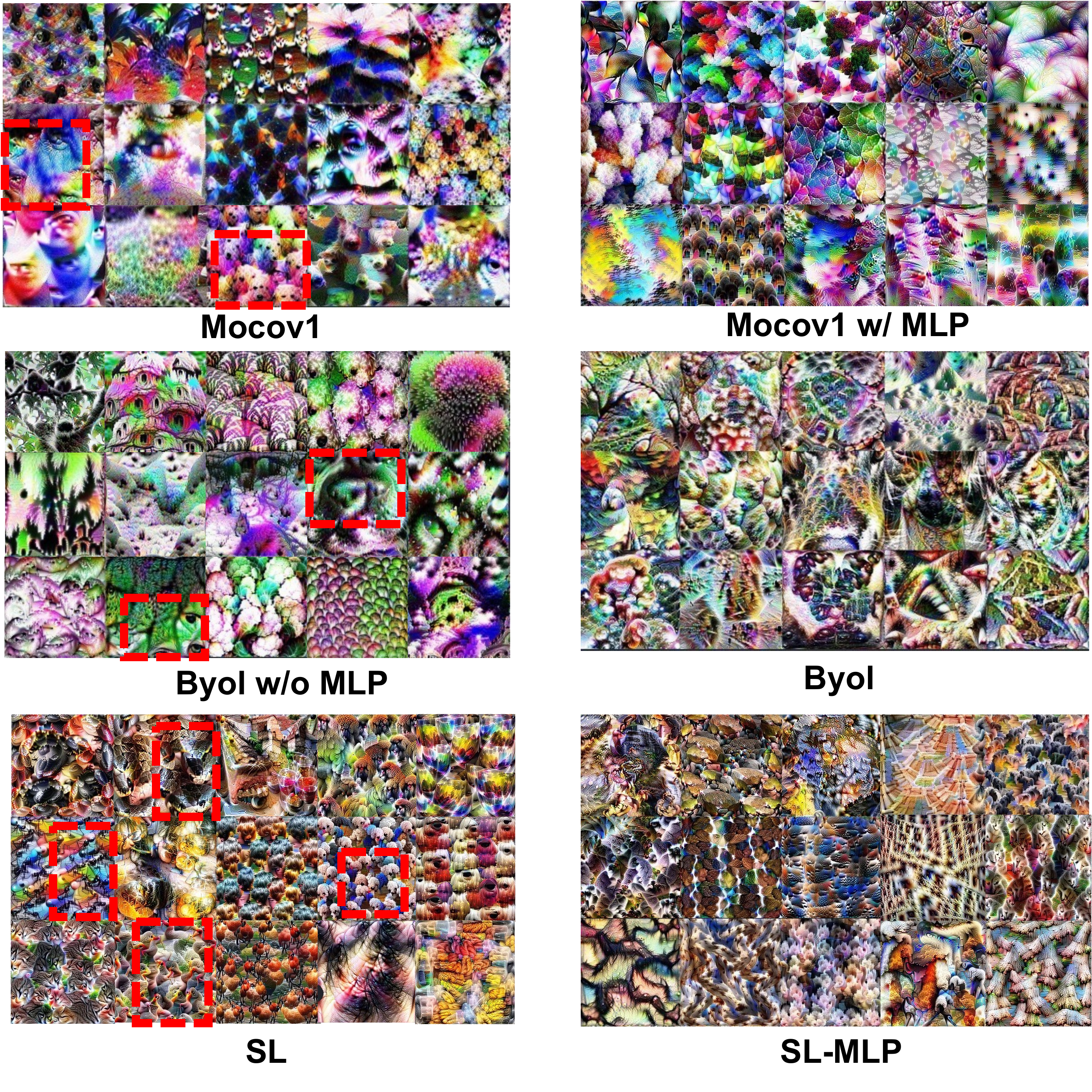}
    \caption{Convolution channels visualization of Mocov1, Mocov1 w/ MLP, Byol w/o MLP, Byol, SL and SL-MLP. Following the method proposed in~\cite{olah2017feature}, we visualize the maximum response of convolution channels in layer 4 of ResNet50 pretrained with different methods.  }
    \label{fig:appe_layer_vis}
\end{figure*}


\section{Detailed Training Setup} \label{appendix:training_setup}
\subsection{Pretraining}
\label{appendix:pretrain}
For SL and SL-MLP, we use the SGD optimizer with a cosine decay learning rate of 0.4 with Nesterov momentum of 0.9 to optimize all the networks and set the batch size to 1024. A 3 epochs warm-up with a starting learninig rate of 0.1 is applied. The weight decay of ResNets, MobileNetv2, EfficientNetb2 is set to $1\times10^{-4}$, $5\times10^{-5}$, $1\times10^{-5}$, respectively. Data augmentations include random-crop (224x224), color-jitter, and random horizontal flip.
For SupCon and SupCon w/o MLP pretraining, we set the temperature parameter to $\tau=0.07$, and queue size to 65596. We use random-crop (224x224), color-jitter, random gray-scale, Gaussian blur, random horizontal flip for pretraining data augmentations.


\subsection{Concept Generalization Task}
In unseen generalization task, we divide ImageNet-1K into two class-exclusive datasets following the hierarchical structure built in WordNet~\cite{miller1998wordnet} - one for pretraining (denoted as pre-D) and the other for evaluation (denoted as eval-D). Eval-D has 348 classes of instrumentality, and pre-D contains 652 classes mostly of organisms.
All the networks are pretrained on pre-D,
and then examined by linear evaluation protocal on eval-D.
As in~\cite{oord2018representation,kolesnikov2019revisiting,chen2020a}, we train a linear classifier with the frozen backbone for 100 epochs. During evaluation, images are resized to 256 pixels, after which $224\times224$ center crop is used. We optimize the cross-entropy loss with SGD optimizer with cosine decay scheduler with Nesterov momentum of 0.9 over 100 epochs, using a batch size of 4096. We finally sweep over 7 learning rate over $\{0.16, 0.48, 1.44, 4.8, 14.4, 48\}$ and report the best accuracy on the test set of eval-D.

\subsection{Transfer to Other Classification Tasks}
\label{appendix:12-domain classification}
Follow the downstream image classification tasks and the evaluation methods mentioned in~\cite{islam2021a}, we use 12 datasets from different domains to evaluate the transferability of different methods, including natural~\cite{mohanty2016using,nilsback2008automated,olsen2019deepweeds}, satellite~\cite{helber2019eurosat,cheng2017remote}, symbolic~\cite{lake2015human,netzer2011Reading}, illustrative~\cite{tian2020kaokore,wang2019learning}, medical~\cite{codella2019skin,wang2017chestx}, and texture~\cite{cimpoi2014describing}. The statistics of datasets are illustrated in Tab.~\ref{tab:appendix-datasets}.

\noindent\textbf{Linear Evaluation.} For fixed-feature linear evaluation, we add a linear layer on the frozen pretrained backbone to train the model on the downstream datasets. A batch normalization layer is added between the backbone and linear layer. All models are trained for 50 epochs with step learning scheduler which decreases the learning rate by 0.1 at epoch 25 and 37. 70\% of the training set is used for training and the rest is used for validation, the models are then trained with
\begin{itemize}
    \item learning rate: 0.001, 0.01, 0.1;
    \item batch size: 32, 128;
    \item weight decay: 0, $1\times10^{-4},1\times10^{-5}$.
\end{itemize}
The optimal hyperparameters are chosen based on the performance on the validation set. The top-1 accuracy is reported as the evaluation metric.

\noindent\textbf{Full Network Finetuning.}
In full network finetuning, the whole pretrained backbone and a linear classifier are trained on the downstream dataset. All models are trained for 50 epochs with step learning scheduler which decreases the learning rate by 0.1 at epoch 25 and 37. A batch normalization layer is added between the backbone and linear layer to make the extracted features comparable among different models. The models are trained with
\begin{itemize}
    \item learning rate: 0.001, 0.01, 0.1;
    \item batch size: 32, 128;
    \item weight decay: 0, $1\times10^{-4},1\times10^{-5}$.
\end{itemize}
The optimal hyperparameters are chosen based on the performance on the validation set.

\noindent\textbf{Few-shot Learning.}
For few-shot learning, following~\cite{tian2020rethinking}, we use a logistic regression layer on the top of the features during meta-testing phase. The implementation from scikit-learn is used for logistic regression. Same as~\cite{islam2021a}, we also provide the mean of 600 randomly sampled tasks as the accuracy.

\begin{table}[htbp]
  \centering
  \small
  \caption{\small{Datasets used for downsteam classification tasks from different domains. Following~\cite{islam2021a}, we divided these datasets into six categories, including satellite, natural, symbolic, medical, illustrative, and texture.}}
    \begin{tabular}{llrrr}
    \toprule
    Category & Dataset & Train Size & Test Size & Classes \\
    \midrule
    \multirow{2}[2]{*}{Satellite} & EuroSAT & 18900 & 8100  & 10 \\
          & Resisc45 & 22005 & 9495  & 45 \\
    \midrule
    \multirow{3}[2]{*}{Natural} & CropDisease & 43456 & 10849 & 38 \\
          & Flowers & 1020  & 6149  & 102 \\
          & DeepWeeds & 12252 & 5257  & 9 \\
    \midrule
    \multirow{2}[2]{*}{Symbolic} & Omniglot & 9226  & 3954  & 1623 \\
          & SVHN  & 73257 & 26032 & 10 \\
    \midrule
    \multirow{2}[2]{*}{Medical} & ISIC  & 7007  & 3008  & 7 \\
          & ChestX & 18090 & 7758  & 7 \\
    \midrule
    \multirow{2}[2]{*}{Illustrative} & Kaokore & 6568  & 821   & 8 \\
          & Sketch & 35000 & 15889 & 1000 \\
    \midrule
    Texture & DTD   & 3760  & 1880  & 47 \\
    \bottomrule
    \end{tabular}%
  \label{tab:appendix-datasets}%
\end{table}%

\begin{table*}[t]
  \centering
\caption{\small 5-ways 5-shots and 20-shots classification performance on 12 downstream datasets in terms of top-1 accuracy. Using the code in~\cite{islam2021a}, we pretrain all models over 300 epochs on ImageNet-1K. The reported accuracy is the mean of 600 randomly sampled tasks. Average results style: \textbf{best}, \underline{second best}.}
  \resizebox{\textwidth}{!}{
    \begin{tabular}{lccccccccccccc}
    \toprule
    Method & ChestX & CropDisease & DeepWeeds & DTD  & EuroSAT & Flowers102 & Kaokore & Omniglot & Resisc45 & Sketch & SVHN  & ISIC & Average \\
    \midrule
    \emph{5-ways 5-shots few-shot classification} \\
    \midrule
    SL             & 25.64 & 89.07 & 54.32 & 78.58 & 82.96 & 93.14 & 46.14 & 92.82 & 84.17 & 87.06 & 38.03 & 41.22 & 67.76\\
    SL-MLP         & 26.89 & 93.45 & 59.08 & 83.04 & 87.16 & 96.88 & 50.77 & 95.73 & 89.00 & 89.84 & 41.96 & 46.76 & \textbf{71.71}\\
    SupCon w/o MLP & 23.62 & 75.64 & 49.34 & 73.04 & 73.90 & 82.16 & 38.10 & 67.87 & 75.18 & 81.01 & 34.92 & 35.16 & 59.16\\
    SupCon         & 26.18 & 94.09 & 59.36 & 85.02 & 87.97 & 96.55 & 51.02 & 94.49 & 89.01 & 89.75 & 41.67 & 43.48 & \underline{71.55}\\
        \midrule
    \emph{5-ways 20-shots few-shot classification} \\
    \midrule
    SL    & 30.05  & 94.15  & 64.54  & 85.74  & 89.13  & 96.63  & 55.65  & 97.17  & 90.34  & 93.12  & 48.09  & 52.06  & 74.72  \\
    SL-MLP & 32.57  & 97.27  & 70.11  & 89.46  & 92.39  & 98.79  & 61.32  & 98.60  & 94.19  & 93.68  & 54.62  & 58.29  & \textbf{78.44}  \\
    SupCon w/o MLP & 26.50  & 84.90  & 57.81  & 80.64  & 82.37  & 89.47  & 46.19  & 83.56  & 83.51  & 88.12  & 44.60  & 44.51  & 67.68  \\
    SupCon & 31.20  & 97.06  & 69.48  & 90.24  & 92.62  & 98.65  & 61.35  & 98.03  & 93.82  & 95.38  & 54.16  & 54.67  & \underline{78.06}  \\

    \bottomrule
    \end{tabular}}%
  \label{tab:appen_few_shots}%
\end{table*}%


\begin{table*}[t]
  \centering
\caption{\small Full-data finetuning classification performance on 12 downstream datasets in terms of top-1 accuracy. Using the code in~\cite{islam2021a}, we pretrain all models over 300 epochs on ImageNet-1K. The reported accuracy is the mean of 600 randomly sampled tasks. Average results style: \textbf{best}, \underline{second best}.}
  \resizebox{\textwidth}{!}{
    \begin{tabular}{lccccccccccccc}
    \toprule
    Method & ChestX & CropDisease & DeepWeeds & DTD  & EuroSAT & Flowers102 & Kaokore & Omniglot & Resisc45 & Sketch & SVHN  & ISIC & Average \\
    \midrule
    \emph{Full-data finetuning} \\
    \midrule
    SL             & 57.71 & 99.87 & 96.88 & 73.78 & 98.60 & 94.31 & 88.80 & 86.37 & 89.55 & 95.90& 78.46 & 97.07 & 88.11\\
    SL-MLP         & 57.98 & 99.88 & 96.90 & 74.26 & 98.77 & 95.12 & 89.16 & 88.81& 90.06 & 96.15 & 79.83 & 97.13 & \underline{88.67}\\
    SupCon w/o MLP & 57.70 & 99.86 & 96.04 & 74.04 & 98.38 & 94.60 & 87.03 & 85.10 & 90.24 & 95.68 & 80.85 & 97.15 & 88.06\\
    SupCon         & 58.61 &99.90 & 96.29 & 75.43& 98.83 & 95.10 & 88.83 & 87.35 & 91.25 & 95.72 & 81.10 & 97.23 & \textbf{88.80}\\

    \bottomrule
    \end{tabular}}%
  \label{tab:appen_finetune}%
\end{table*}%

\subsection{Object Detection on COCO}
For object detection, we train Mask-RCNN~\cite{he2017mask} (R50-FPN) on COCO 2017 train split and report results on the val split. We use a learning rate of 0.001 and keep the other parameters the same as in the $1\times$ schedule in detectron2~\cite{wu2019detectron2}.

\section{More Results}
\subsection{Few-shot Recognition Results}
Using the code provided by~\cite{islam2021a}, we pretrain all models over 300 epochs with a cosine decay learning scheduler on ImageNet-1K, and then testing on 12 downstream datasets (shown in~Tab.\ref{tab:appendix-datasets}). We provide 5-ways 5-shots and 20-shots results in Tab.~\ref{tab:appen_few_shots}. All reported  accuracy is the mean of 600 randomly sampled tasks. Comparing average results among different methods, we observe that supervised pretraining methods with the MLP projector, \emph{i.e.} SL-MLP and SupCon, outperform their no MLP counterparts, \emph{i.e.} SL and SupCon w/o MLP, on both 5-ways 5-shots and 20-shots few-shot classification tasks.

\subsection{Full-data Finetuning Results}
We also provide full-sample results of 12-domains transfer task in Tab.~\ref{tab:appen_finetune}, SL-MLP still gets a \textbf{+0.56\%} accuracy gain. These consistent results show that adding an MLP on SL has large improvement on linear evaluation and observable improvement on fine-tuning (though relatively smaller).

\subsection{Original Concept Generalization Task}

\begin{table}[tbp]
  \centering
  \caption{Original concept generalization task~\cite{sariyildiz2021concept} results. SL, SL-MLP, and Byol are all pretrained on ImageNet-1K with 300 epochs. Following~\cite{sariyildiz2021concept}, L1/L2/L3/L4/L5 represent five ImageNet-1K-sized datasets of increasing semantic distance from IN-1K as concept generalization levels.}
    \begin{tabular}{lccccccc}
    \toprule
    \multicolumn{1}{l}{} & Epoch & \multicolumn{1}{c}{In-1K} & \multicolumn{1}{c}{L1} & \multicolumn{1}{c}{L2} & \multicolumn{1}{c}{L3} & \multicolumn{1}{c}{L4} & \multicolumn{1}{c}{L5} \\
    \midrule
    SL    & 300   & \textbf{77.0}    & 66.2  & 60.1  & 56.1  & 54.7  & 48.3 \\
    Byol  & 300   & 71.7  & 68.2  & 64.4  & 60    & 58.7  & 52.9 \\
    SL-MLP & 300   & 75.6  & \textbf{70.4}  & \textbf{66.2}  & \textbf{61.8}  & \textbf{60.8}  & \textbf{54.4} \\
    \bottomrule
    
    \end{tabular}%
    \label{tab:appen_concept_generalization_task}
\end{table}

We pretrain SL, SL-MLP, and Byol on ImageNet-1K with 300 epochs and use the code provided by~\cite{sariyildiz2021concept} to evaluate their transferability on five ImageNet-1K-sized datasets of increasing semantic distance from IN-1K. Results are summarized in Tab.~\ref{tab:appen_concept_generalization_task}. SL-MLP is better than SL and Byol, and the improvement increases when the semantic distance increases from L1 (\textbf{+4.2}\%) to L5 (\textbf{+6.1}\%).

\subsection{Ablation on Hidden Units and Output Dimensions}
On concept generalization task, we also explore whether hidden units and output dimensions of the added MLP projector influence the final transferability.  We pretrain SL-MLP on pre-D over 100 epochs using various hidden units and output dimensions of the added MLP projector, and report the evaluation results on eval-D (illustrated in~Fig.\ref{fig:appen_ablation_hidden_units_output_dimens}). We observe that, different from other unsupervised pretraining methods, \eg BYOL and SimCLR, where the output dimension of the MLP projector have considerable impacts on transferability,  the hidden units and output dimensions of the added MLP projector has little influence on the performance of SL-MLP. 

\begin{figure*}
    \centering
    \subfloat[]
    {\includegraphics[width=.42\linewidth]{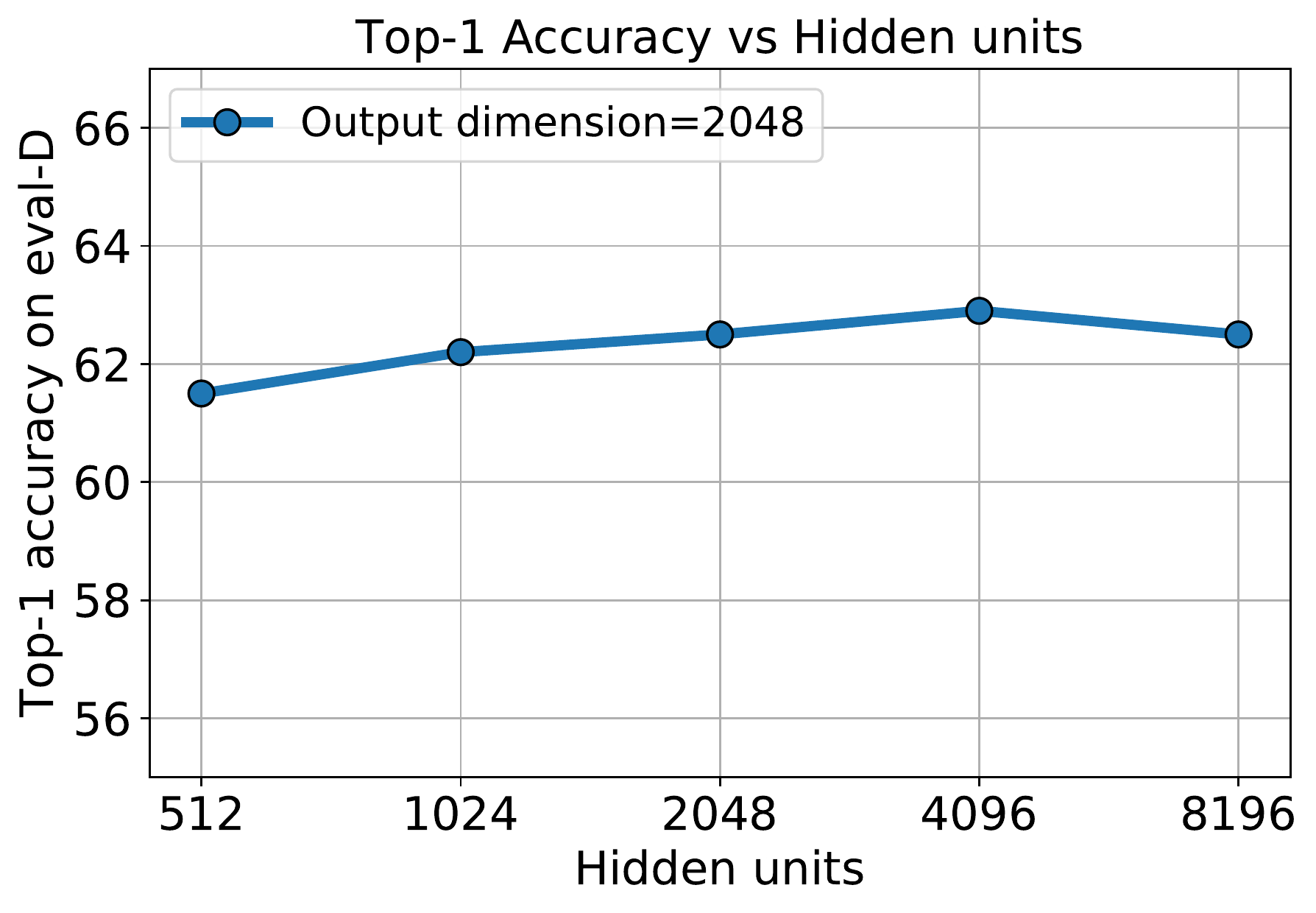}\vspace{-0.05cm}}\hfill
    \subfloat[]
    {\includegraphics[width=.42\linewidth]{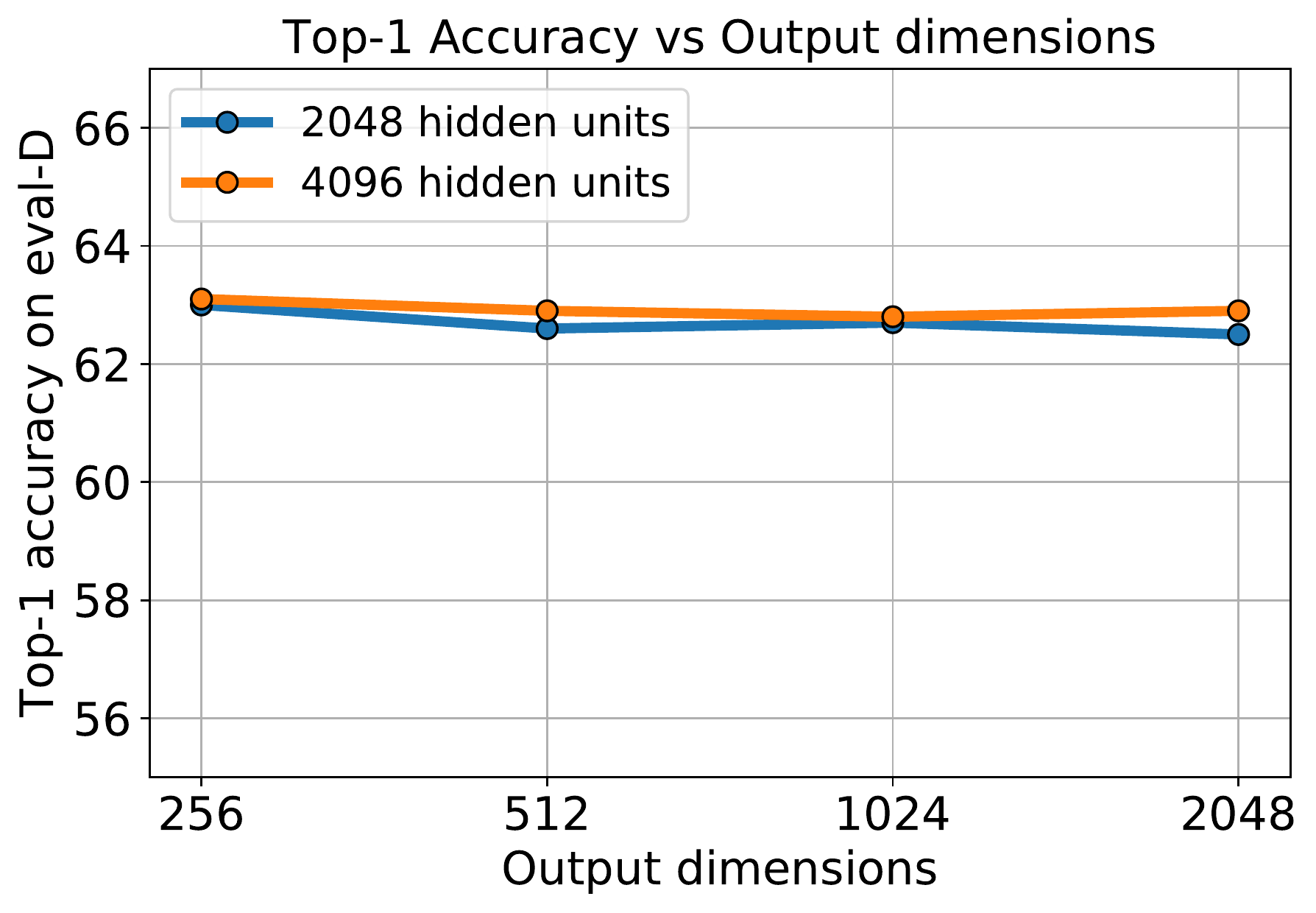}\vspace{-0.05cm}}\hfill
    \caption{\small{\textbf{(a)} Top-1 accuracy on eval-D as a function of the number of hidden units of the added MLP projector. \textbf{(b)} Top-1 accuracy on eval-D as a function of output dimension of the added MLP projector. We pretrain all the models on pre-D over 100 epochs and then evaluate on eval-D. Both hidden units and output dimensions show slight influence on the improved transferability.}}
    \label{fig:appen_ablation_hidden_units_output_dimens}
\end{figure*}

\begin{table*}[t]
    \small
  \centering
  \caption{\small{Linear evaluation results and top-1 accuracy during pretraining on SL and SL-MLP. We remove the MLP in SL-MLP for linear evaluation, only the fixed backbones of SL and SL-MLP are used. For top-1 accuracy during pretraining, accuracy of the whole SL-MLP is reported.  }}
    \begin{tabular}{ccccc}
    \toprule
    \multirow{2}[4]{*}{Epochs} & \multicolumn{2}{c}{Top-1 accuracy during pretraining} & \multicolumn{2}{c}{Linear evaluation accuracy of fixed backbones} \\
\cmidrule{2-5}          & SL    & SL-MLP & SL    & SL-MLP \\
    \midrule
    20     & 59.1  & 51.5   & 70.0  & 66.0\\
    40     & 64.0  & 61.2   & 71.6  & 69.1\\
    60     & 69.4  & 69.2   & 74.8  & 72.8\\
    80   & 76.6  & 76.7 & 78.5  & 75.8   \\
    100   & 80.8  & 80.2 & 80.8  & 78.2    \\
    \bottomrule
    \end{tabular}%
  \label{tab:appendix-pretrain-results}%
\end{table*}%

\begin{figure}[!t]
    \centering
    \includegraphics[width=0.95\textwidth]{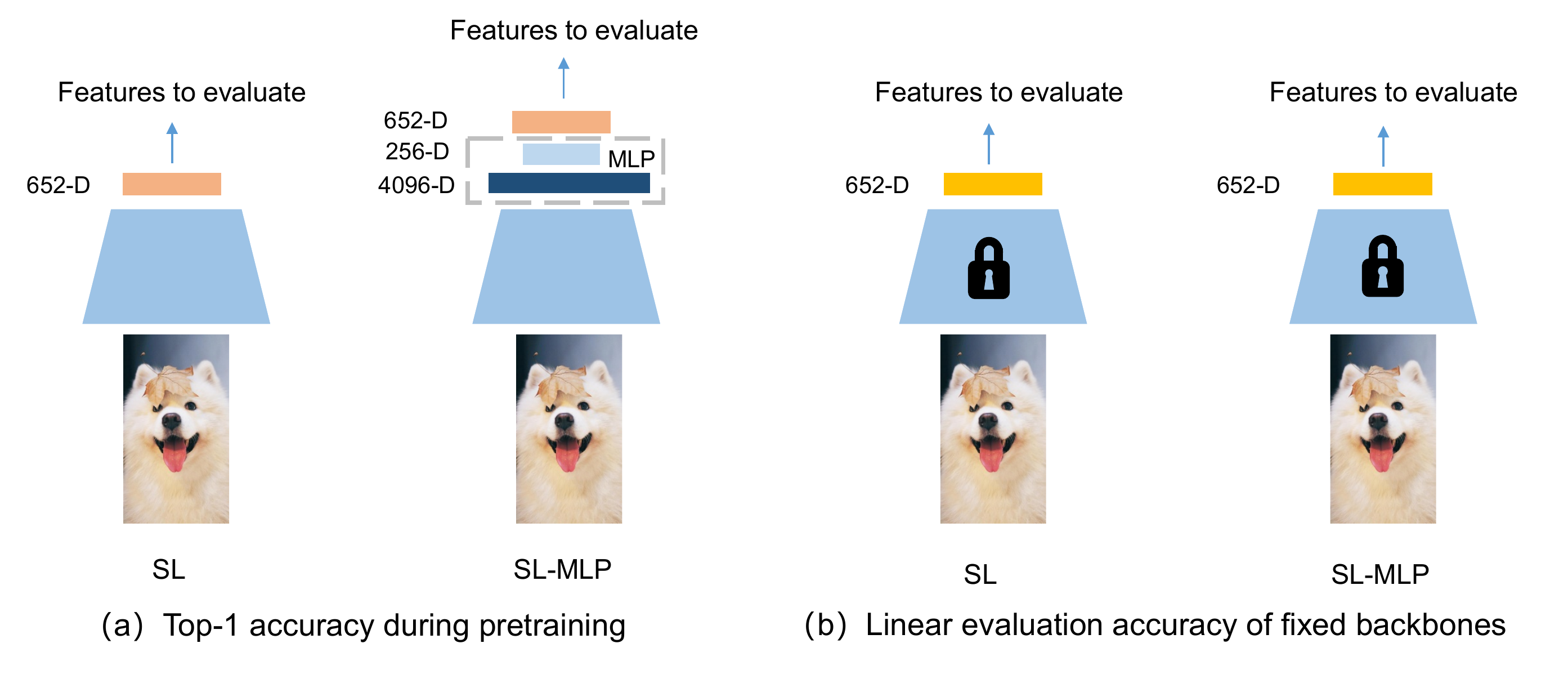}
    \caption{\small{Evaluation of features extracted by SL and SL-MLP. \textbf{(a)}}: During pretraining, features after the classifier is used to evaluate the accuracy on pre-D. \textbf{(b)}: After pretraining, we use the fixed backbones from different epochs to evaluate the performance of SL and SL-MLP.  }
    \label{fig:appendix feature to eval}
\end{figure}

\section{Pretrain Results on pre-D}
We also provide the top-1 accuracy of SL-MLP on pre-D in Tab.~\ref{tab:appendix-pretrain-results}. We remove the MLP in SL-MLP for linear evaluation on pre-D, only the fixed backbones of SL and SL-MLP are used to train new classifiers over 100 epochs. We also report top-1 accuracy during pretraining in which accuracy of the whole SL-MLP is reported. Which features are used to evaluate these two metrics are illustrated in Fig.~\ref{fig:appendix feature to eval}.
As backbones and classifiers are jointly trained during pretraining, classifiers are not well optimized at small pretraining epochs. Thus, models always achieve better performance on linear evaluation at small pretraining epochs because linear evaluation provides more epochs for networks to optimize better classifiers on fixed backbones. For SL, two evaluation methods display the same result at epoch 100, as they have all trained well-optimized classifiers.

Note that SL-MLP shows slight $-2.6\%$ performance drop (80.8\% to 78.2\%) on linear evaluation when SL and SL-MLP have all been pretrained over 100 epochs, which achieves closer performance gap than Exemplar-v2~\cite{zhao2021what} when compared with SL.
Besides, as SL-MLP only adds an MLP projector before the classifier, the whole SL-MLP shows almost the same performance of SL on top-1 accuracy during pretraining at epoch 100. 

\end{onecolumn}
\end{document}